\documentclass[11pt]{article}
\usepackage[hmargin=1in,vmargin=1in]{geometry}
\usepackage[parfill]{parskip}
\usepackage{graphicx}
\usepackage{amssymb}
\usepackage{epstopdf}
\usepackage{amsmath}
\usepackage{eucal}
\usepackage{algorithm}
\usepackage{algorithmic}
\usepackage[usenames]{color}
\usepackage{multicol} 
\usepackage{mdwlist}
\usepackage{url}
\usepackage{todonotes}
\usepackage{multirow}
\usepackage{multicol}

\title{Dimension Reduction Using Rule Ensemble Machine Learning Methods: A Numerical Study of Three Ensemble Methods}
\author{Orianna DeMasi\footnote{ODemasi@lbl.gov}, Juan Meza\footnote{JCMeza@lbl.gov}, David H. Bailey\footnote{DHBailey@lbl.gov}\\Lawrence Berkeley National Laboratory\\1 Cyclotron Road\\Berkeley, CA 94720 }
\date{ }
\begin{document}
\maketitle

\begin{abstract}
Ensemble methods for supervised machine learning have become popular due to their ability to accurately predict class labels with groups of simple, lightweight ``base learners." While ensembles offer computationally efficient models that have good predictive capability they tend to be large and offer little insight into the patterns or structure in a dataset. We consider an ensemble technique that returns a model of ranked rules. The model accurately predicts class labels and has the advantage of indicating which parameter constraints are most useful for predicting those labels. An example of the rule ensemble method successfully ranking rules and selecting attributes is given with a dataset containing images of potential supernovas where the number of necessary features is reduced from 39 to 21. We also compare the rule ensemble method on a set of multi-class problems with boosting and bagging, which are two well known ensemble techniques that use decision trees as base learners, but do not have a rule ranking scheme. \let\thefootnote\relax\footnotetext{This research was supported in part by the Director, Office of Computational and Technology Research, Division of Mathematical, Information, and Computational Sciences of the U.S. Department of Energy, under contract number DE-AC02-05CH11231}
\end{abstract}
%
\section{Introduction}
Machine learning algorithms are popular tools for classifying observations. These algorithms can attain high classification accuracy for datasets from a wide variety of applications and with complex behavior.  In addition, through automated parameter tuning, it is possible to grow powerful models that can successfully predict class affiliations of future observations. A disadvantage, however, is that models can become overly complicated and, as a result, hard to interpret and expensive to evaluate for large datasets. Ideally we would like to generate models that are quick to build, cheap to evaluate, and that give users insight into the data, similar to how the size of coefficients in a linear regression model can be used to understand attribute-response relationships and dependencies. \\

Ensemble methods are a class of machine learning algorithms that develop simple and fast algorithms by combining many elementary models, called base learners, into a larger model. The larger model captures more behavior than each base learner captures by itself and so collectively the base learners can model the population and accurately predict class labels \cite{friedman03}. Classical decision tree ensemble methods, such as bagging and boosting, are well known and have been tested and refined on many datasets \cite{banfield04, dietterich00, quinlan96}. In one such study, Banfield et al. \cite{banfield04} studied the accuracy of boosting and bagging on a variety of public datasets and found that in general neither bagging nor boosting was a statistically significantly stronger method. \\

In this paper, we modify, extend, and test an implementation \cite{meza09} of the rule ensemble method proposed by Friedman and Popescu \cite{friedman05} for binary classification with bagging and with boosting. The Friedman and Popescu rule ensemble method is attractive, as it combines the rule weighting or variable importance that regression provides with the quick decision tree methods and collective decision making of many simple base learners. The method builds rules, that take the form of products of indicator functions defined on hypercubes in parameter space. The rules are fit by growing decision trees, as each inner node of a tree takes the desired form of a rule. The method then performs a penalized regression to combine the rules into a sparse model. The entire method resembles a linear regression model, but with different terms. Many ensemble methods provide little insight into what variables are important to the behavior of the system, but by combining the rules with regression, the rule ensemble method prunes rules of little utility and ranks remaining rules in order of importance.\\

We also modified the rule ensemble method to use various coefficient solving methods on a set of binary and multi-class problems.  Previous implementations of this algorithm are either currently unavailable~\cite{rulefit} or have not been fully tested on a wide set of problems~\cite{root}. We extended the rule ensemble method to multiple class classification problems with one versus all classification \cite{rifkin04} and tested it on classical machine learning datasets from the UC Irvine machine learning repository \cite{uci}. These datasets were chosen because they have been used to test previous tree ensembles \cite{banfield04, dietterich00, hastie01, quinlan96, witten05} and countless other machine learning algorithms. Finally, we  look at different methods that can be used to solve for the coefficients and show how one can use the rule ensemble method to reduce the dimension of a problem. We give an example of identifying important attributes in a large scientific dataset by applying our techniques to a set of images of potential supernova \cite{nugent}.

\subsection{Overview of Rule Ensemble Method}
Suppose we are given a set of data points $\{\mathbf{x}_i, y_i\}_{i = 1}^N$, where $\mathbf{x}_i$ denotes the $i$th observation, with label $y_i$. Each of the observations, $\mathbf{x} \in \mathbb{R}^\mathcal{K}$, has $\mathcal{K}$ attributes or feature values that we measure for each observation. The matrix $\mathbf{X}$ will denote the entire set of all $\mathbf{x}_i$'s. The $j$th feature of the $i$th observation is the scalar $x_{ij}$. Our goal then is to be able to predict what class $y$ a future unlabeled observation $\mathbf{x}$ belongs to. The method below focuses specifically on the binary decision problem where $y$ can be one of only two classes $\{-1, +1\}$. To classify observations we seek to construct a function $F(\mathbf{x})$ that maps an observation $\mathbf{x}$ to an output variable $\hat{y} = F(\mathbf{x})$ that predicts the true label $y$. \\

Define the risk of using any function that maps observations to labels as 
\begin{equation} R(F) = E_{\mathbf{x},y} L(y,F(\mathbf{x})), \label{eq:risk} \end{equation}
where $E_{\mathbf{x},y}$ is the expectation operator. $L(y,\hat{y})$ is a chosen loss function that defines the cost of predicting a label $\hat{y}$ for an observation when the true label is $y$. While various loss functions have been developed, in practice we will use the ramp loss function as it is particularly well suited to the binary classification problem we consider \cite{friedman03, friedman05}. Within this framework we seek to find a function, $ F^*(\mathbf{x})$, that minimizes the risk over all such functions
$$F^*(\mathbf{x}) = \underset{F}{\operatorname{argmin}}\ E_{\mathbf{x},y} L(y,F(\mathbf{x})).$$

The optimal $F^*(\mathbf{x})$ is defined on the entire population. However, we only have a training sample of observed data $\mathcal{S} =\{\mathbf{x}_i, y_i\}_{i = 1}^N$ so we will construct an approximation $\hat{F}(\mathbf{x})$ to $F^*(\mathbf{x})$ that minimizes the expected loss on this training set. We assume that the model $\hat{F}(\mathbf{x})$ has the form of a linear combination of $K$ base learners $\{f_k(\mathbf{x})\}_{k = 1}^K$: 
\begin{equation}
\hat{F}(\mathbf{x}; \mathbf{a}) = a_0 + \sum_{k = 1}^K a_kf_k(\mathbf{x}). 
\end{equation}

The next step is to find coefficients $\mathbf{a} = \{a_1, a_2, \ldots, a_K \}$ that minimize the risk (\ref{eq:risk}). Like $F^*(\mathbf{x})$, the risk is defined over the entire population, so we will use the approximation $\mathbf{a}^*$  that minimizes the risk over the given sample set of observations $\mathcal{S}$. In particular, we take $\mathbf{a}^*$ to be the solution of 
\begin{eqnarray}
\mathbf{a}^* &&=\underset{\{\mathbf{a}\}}{\operatorname{argmin}}\ E_{\mathcal{S}} L (y,F(\mathbf{x}; \mathbf{a})),\\
&&=\underset{\{\mathbf{a}\}}{\operatorname{argmin}} \frac{1}{N}\sum_{i = 1}^N L(y_i, F(\mathbf{x}_i; \mathbf{a})),\\
&&= \underset{\{\mathbf{a}\}}{\operatorname{argmin}} \frac{1}{N} \sum_{i = 1}^N L \left( y_i, a_0 + \sum_{k = 1}^K a_k f_k(\mathbf{x}_i)\right).\label{eq:astar}
\end{eqnarray}

If the loss function, $L$, is taken to be the mean squared error then this is simply a linear regression problem. \\

In many cases, a solution to equation (\ref{eq:astar}) is not be the best for constructing a sparse interpretable model or a predictive model that is not overfit to the training data. Instead, one would like to have a solution that has as few components as possible. To achieve a sparse solution, a penalty term can be included that prevents less influential terms from entering the model. Here, we use the $L^1$ (lasso \cite{tibshirani96}) penalty and the approximation $\hat{\mathbf{a}}$, which is the solution to the penalized problem 
\begin{equation}
\mathbf{\hat{a}} = \arg \min_{\{\mathbf{a}\}} \sum_{i = 1}^N L\left(y_i, a_0 + \sum_{k = 1}^K a_k f_k(\mathbf{x}_i)\right) + \lambda \sum_{k = 1}^K |a_k|.
\label{eq:regularizedreg} 
\end{equation}
The impact of the penalty is controlled by the parameter $\lambda \geq 0$. This penalized problem has received a great deal of attention \cite{donoho95, friedman09, hale07} and enables both estimation of the coefficients as well as coefficient selection.\\

This section provided a brief introduction to the methods used in this study and that were developed by Friedman and Popescu \cite{friedman03, friedman04, friedman05}. Other papers provide more details and and justification of the rule ensemble method \cite{friedman03, friedman05} as well as the method that is used to assemble the rules in the latter part of the algorithm \cite{friedman04}. Additional sources also provide more details for the other algorithms that we employed to compute the coefficients \cite{friedman09, hale07, berg08}.

In section \ref{sec:baselearners}, we will discuss how to build base learners $f_k$. Section \ref{sec:Pathbuild} will provide more details on the regression method used to solve equation (\ref{eq:regularizedreg}). Sections \ref{sec:multi-class}-\ref{sec:binary-class} will present computational results comparing the rule ensemble method with other ensemble methods.

\section{Base Learners}\label{sec:baselearners}
The base learners $f_k$ in equation (\ref{eq:Fhat}) can be of many different forms. Decision trees, which have been used alone as classification models, have been used as base learners in ensemble methods such as bagging, random forests, and boosting. Decision trees are a natural choice to use for a learner, as many small trees (meaning each tree has few leaves) can be built quickly and then combined into a larger model. The bagging method grows many trees, then combines them with equal weights \cite{breiman96}. Boosting is more sophisticated as it tries to build the rules in an intelligent manner, but it still gives each tree an equal weight in the ensemble \cite{freund96}.

\subsection{Using Rules as Base Learners}
In the rule ensemble method,  simple rules denoted by $r_k$ are used as the base learners and take the form 
\begin{equation}r_k(\mathbf{x}_i) = \prod_{j}I(x_{ij} \in p_{kj}), \label{eq:ruleform} \end{equation}
where $I(x_{ij} \in p_{kj}) $ is an indicator function. The indicator function evaluates to $1$ if the observed attribute value $\mathbf{x}_{ij}$ is in the parameter space defined by $p_{kj}$, and $0$ if the observation is not in that space. Each $p_{kj}$ is a constraint that the $k$th rule assigns to the $j$th attribute.  For convenience we will denote $\mathbf{p}_k = (p_{k1}, \dots)$ 
to be the vector of parameter constraints that an observation must meet to have the $k$th rule to evaluate to $1$. Note that a given rule can have multiple constraints on a single attribute, as well as a different number of constraints (indicator functions) than other rules. To emphasize that each rule is defined by a set of parameters we can write $r_k(\mathbf{x}_i) = r_k(\mathbf{x}_i; \mathbf{p}_k)$.\\ 

To fit a model we need to generate rules by computing parameter sets $\{\mathbf{p}_k\}_{k= 1} ^K$.  In this study, we will use decision trees to generate rules, where each internal and terminal node (not the root node) of a decision tree takes the form of a simple rule defined by (\ref{eq:ruleform}). Having $r_k(\mathbf{x}_i; \mathbf{p}_k) = 1$ means that the $k$th rule is obeyed by the $i$th observation and that it was sorted into the $k$th node of the decision tree that generated the rule. 

\subsection{Tree Construction - Rule Generation}
\begin{figure}
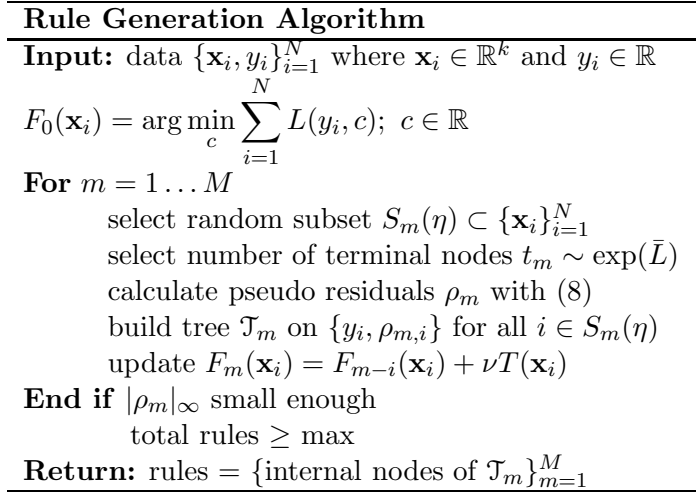
 \centering
\begin{tabular}{l}\hline 
\textbf{Rule Generation Algorithm} \\\hline
\textbf{Input:} data $ \{\mathbf{x}_i, y_i\}_{i = 1}^N$ where $\mathbf{x}_i \in \mathbb{R}^k$ and $y_i \in \mathbb{R}$\\
$F_0(\mathbf{x}_i) = \arg \displaystyle \min_{c} \sum_{i = 1}^N L(y_i,c); \text{ } c\in \mathbb{R}$\\
\textbf{For} $ m = 1 \dots M$\\
\hspace{10mm} select random subset $S_m(\eta) \subset \{\mathbf{x}_i\}_{i = 1}^N$\\
\hspace{10mm} select number of terminal nodes $t_m \sim \exp(\bar{L})$ \\
\hspace{10mm} calculate pseudo residuals $\mathbf{\rho}_m$ with (\ref{eq:pseudoresiduals})\\
\hspace{10mm} build tree $\mathcal{T}_m$ on $\{y_i, \rho_{m,i} \}$ for all $i \in S_m(\eta)$\\
\hspace{10mm} update $F_m(\mathbf{x}_i) = F_{m-i}(\mathbf{x}_i) + \nu T(\mathbf{x}_i)$\\
\textbf{End if} $|\rho_m|_\infty$ small enough\\	
\hspace{13mm} total rules $\geq$ max \\
\textbf{Return:} rules = $ \{ \text{internal nodes of } \mathcal{T}_m\}_{m = 1}^M $\\ \hline
 \end{tabular}
 \caption{Outline of how to generate rules}
\label{tab:Rule Generation Algorithm}
\end{figure}

Decision trees are built using the CART (Classification and Regression Trees) algorithm \cite{breiman1998classification}, which is summarized Table \ref{tab:Rule Generation Algorithm} and outlined below. We let 
$$ \mathcal{T}_m = \{r^m_j\}_{j = 1} ^{2(t_m-1)}$$
 denote the set of rules contained in the $m$th tree which has $t_m$ terminal nodes. Let
 $$T_m(\mathbf{x}_i) = \displaystyle \sum_{j = 1}^{2(t_m-1)} r^m_j(\mathbf{x}_i, p^m_j))$$ 
 denote the prediction that the $m$th tree makes for observation $\mathbf{x}_i$; it is the evaluation of the rules in $\mathcal{T}_m$ on $\mathbf{x}_i$. 

Each tree is built on a random subset of observations $S_m(\eta) \subset \{\mathbf{x}_i, y_i\}_{i = 1}^N$, as training on the entire dataset can be expensive as well as overfit the tree. $\eta$ is a parameter that controls the diversity of the rules by defining the number of observations chosen to be in the subset $S_m(\eta)$. 
As subset size $\eta$ decreases, diversity increases with potentially less global behavior getting extracted. Diversity between the trees can also be increased by varying the final size of each tree. Clearly larger trees include more precise rules defining terminal nodes and thus are inclined to overtrain, but confining the size of a tree too strictly can prevent it from capturing more subtle behavior within the dataset. To avoid under or overfitting, we grow each tree until it has $t_m$ terminal nodes, where $t_m$ is drawn from an exponential distribution. The distribution has mean $\bar{L}$, which does have to be selected {\em a priori}. The size of a tree is determined by growing each branch until no further nodes can be split because one of the following termination conditions has been met:
\begin{enumerate*}
\item The number of observations in a terminal node is less than some selected cutoff,
\item The impurity of a node is less than a selected cut off,
\item The total number of nodes in the tree is greater than $t_m$.
\end{enumerate*}

The splitting attribute and value is chosen as the split that minimizes the sum of the impurities (variance of the node) of the two child nodes if that split were taken. For each split only a random sample of attributes are considered in order to both increase the diversity of learners and decrease training time for huge datasets.

\subsection{Gradient Boosting}
To avoid simply regrowing overlapping rules, with no further predictive capability, we use gradient boosting to intelligently generate diverse rules. With gradient boosting, each tree is trained on the pseudo residuals $\mathbf{\rho}_m$ of the risk function evaluated on the test set rather than training directly on the data \cite{hastie01}. The $i$th element of the pseudo residual vector in the $m$th iteration is given by
\begin{equation}\label{eq:pseudoresiduals}
\mathbf{\rho}_{mi} = - \bigg[ \frac{\partial L(y_i, F(\mathbf{x}_i))}{\partial F(\mathbf{x}_i)} \bigg]_{ F(\mathbf{x}_i) = F_{m-1}(\mathbf{x}_i)} 
\end{equation}
for all $\mathbf{x}_i \in S(\eta)_m$. Each $\mathbf{\rho}_m$ is a vector with as many entries as there are observations in the subsample $S_m(\eta)$ on which it is evaluated. $F_m(\mathbf{x})$ is the memory function at the $m$th iteration. It gives a label prediction based on all the previous learners (trees) that were built. Note that $F_m(\mathbf{x})$ is an intermediate model of trees that is used in rule generation, while $F(\mathbf{x})$ is the final prediction model that has rules as linear terms. Training on the pseudo residuals allows one to account for what the previous trees were unable to capture. This method is similar to the method of regressing on residuals in multidimensional linear regression. Using pseudo residuals also provides another termination condition. If the pseudo residuals shrink below a chosen value, enough behavior has been captured and no further rules are generated. A shrinkage parameter, $0 \leq \nu \leq 1$, controls the dependency of the prediction on the previously built learners. Using $\nu = 0$ results in no dependence on past calculations, so that the next rule is built directly on the data labels and have had no part of the labeled value ``accounted for" by dependence on previous calculations. 

\section{Weighting Rules}\label{sec:Pathbuild}
To combine the rules into a linear model, we need to approximate the coefficients $\hat{\mathbf{a}}$ defined in equation (\ref{eq:regularizedreg}). Here we implement a method that approximates $\hat{\mathbf{a}}$ with an accelerated gradient descent method developed by Friedman and Popescu \cite{friedman04} and summarized in Figure \ref{fig:pathbuild pseudocode}. We will refer to this method as {\sc Pathbuild}, as it does not solve (\ref{eq:regularizedreg}) explicitly, but rather constructs $\hat{\mathbf{a}}$ by starting with a null solution and then incrementing along a constrained gradient descent path, distinguished by a parameter $\tau$. Alternative algorithms for approximating $\hat{\mathbf{a}}$ will be discussed and compared later.\\

We would like find a value for the lasso penalty that yields the sparsest solution to (\ref{eq:regularizedreg}) while maintaining a model with high accuracy. We initialize the coefficients to $0$ and find the constant intercept $a_0$ by the value that minimizes
$$ a_0 = \arg \min_{\alpha} \sum_{i = 1}^N L(y_i, \alpha).$$
This may be better understood by considering that $\alpha$ will be the mean of $(y_1, \dots, y_N)$ when the loss is mean squared error. We approximate $\hat{\mathbf{a}}$ iteratively and calculate the $l+1$st iteration, $\mathbf{a}^{l+1}$, by taking
$$g_k(\mathbf{X}; \mathbf{a}) = \displaystyle\frac{\partial}{\partial a_k} \frac{1}{N} \sum_{i = 1}^N L(y_i, F(\mathbf{x}_i;\mathbf{a})),$$
$$k^* = \{k : g_k(\mathbf{X}; \mathbf{a}^{\ell}) = ||g_k(\mathbf{X}; \mathbf{a}^{\ell})||_\infty \}.$$
We update the coefficients by 
\begin{equation}\label{eq:aupdate}
a_k^{\ell+1}= \begin{cases} 
a_{k}^\ell, & \mathrm{if}\, k = 1..K, k \notin k^* \\
a_{k}^\ell+ \delta g_k(\mathbf{X}; \mathbf{a}^{\ell}) & \mathrm{if}\, k \in k^*,\\
\end{cases}
\end{equation}
where $g_k(\mathbf{X}; \mathbf{a}^{\ell}) $ is the gradient of $F(\mathbf{x})$ calculated with the $\ell$th iteration and evaluated on the entire dataset. The scaling parameter $\delta>0$ can be set constant or chosen at each step in a clever manner. Note that in equation (\ref{eq:aupdate}) only a single component of the coefficient vector is updated at any iteration and thus only a single rule is able to enter the model at an iteration. The method only progresses in the direction of rules which have a large effect on the predictive capability and avoids steps that are of trivial effect. This condition may be relaxed by incrementing all of the components of $\mathbf{a}$ that have a sufficiently large gradient 
$$k^* = \{k : g_k(\mathbf{x}; \mathbf{a}^{\ell}) \geq \tau ||g_k(\mathbf{x}; \mathbf{a}^{\ell})||_\infty \}.$$
The parameter $\tau \in [0,1]$ controls how large a component of the gradient must be relative to the largest component in order for a coefficient to be updated. Computing the gradient is expensive, but reorganizations and intelligent approximations to accelerate the computation are presented for three different loss functions in the appendix \cite{friedman04}. The tricks used for this ``fast" method are most effective for ramp loss and make {\sc Pathbuild} a particularly attractive method. \\

In sections \ref{sec:glmnet}-\ref{sec:fpc} we will compare {\sc Pathbuild} with three different algorithms that can be used to solve for the coefficients. Each algorithm uses a slightly different formulation of the problem defined in equation (\ref{eq:regularizedreg}) and a different technique to encourage a sparse solution that also has little risk. The three algorithms also use mean squared error to define loss rather than the ramp loss function that we use in {\sc Pathbuild}.


\begin{figure} \centering
\begin{tabular}{l}\hline 
\textbf{{\sc Pathbuild}: Gradient Regularized Descent Algorithm} \\\hline
set constant $a_0 = \displaystyle \min_{\alpha} \sum_{i = 1}^N L(y_i, \alpha)$\\
$a^0_k = 0,\text{ } k = 1, \dots, K$\\
\textbf{For} $\ell = 1, \dots,$ max iterations\\
\hspace{10mm} $g_k(\mathbf{x}; \mathbf{a}^{\ell}) = \displaystyle\frac{\partial}{\partial a_k} \frac{1}{N} \sum_{i = 1}^N L(y_i, F(\mathbf{x}_i; \mathbf{a}^\ell))$\\
\hspace{10mm} $a_k^{\ell+1}= \begin{cases} 
a_{k}^\ell, & \mathrm{if}\, g_k(\mathbf{x}; \mathbf{a}^{\ell}) \geq \displaystyle |g_k^\ell|_\infty\\
a_{k}^\ell+ \delta g_k(\mathbf{x}; \mathbf{a}^{\ell}) & \mathrm{otherwise}
\end{cases}$\\
\textbf{Stop if} risk increased \\
\hspace{10mm} $l > $ max iterations\\
\hspace{10mm} gradient $<$ tolerance\\ \hline

 \end{tabular}
\caption{Outline of {\sc Pathbuild} method}\label{fig:pathbuild pseudocode}
\end{figure}


\section{Datasets and Methods for Experiments} \label{sec:supernova_test}
To test the behavior of the rule ensemble method on a binary classification problem, we used a dataset of images taken by a telescope \cite{nugent, 2009PASP..121.1395L, 2009PASP..121.1334R}, the goal being to identify potential supernovas. The initial data had three images for each observation. Those images were processed to yield 39 statistics for each observation that described the distribution and color of individual pixels within the original three images. These statistics became the attributes for the dataset and the observations were labeled with +1, -1 if they were or were not, respectively, an image of a supernova-like object. The dataset contains a total of 5,000 positive and 19,988 negative observations.
To test how the rule ensemble works on the binary classification problem, we use a procedure that first randomly selects 2,500 positive observations and 2,500 negative observations for a training set, and then uses the remaining data for the testing set. This selection process is repeated 10 times for cross-validation. False positive and false negative error rates were used to assess the accuracy of the methods in addition to the overall error rate. The false positive rate is the ratio of observations misclassified as positive to the total number of negative observations in the test set, while the false negative rate is the ratio of observations misclassified as negative to the number of positive observations in the test set. The overall error rate is the ratio of observations misclassified to the total number of observations in the test set. The experiments show the effect of the rule complexity (tree depth), number of rules available (tree size), and $\tau$ thresholding in {\sc Pathbuild} on the accuracy of the method. We also consider the effect of substituting different coefficient solvers in place of {\sc Pathbuild}.\\

To assess the overall utility of the rule ensemble we extend our numerical experiments to multi-class problems, which are described in section \ref{sec:multi-class}. We compare the rule ensemble with classical bagging and boosting methods by testing all three algorithms on 10 datasets from the UC Irvine Machine Learning Data Repository \cite{uci} with five 2-fold cross-validation tests. A 2-fold cross-validation test is similar to the method described above except that the dataset is split into equally sized subsets with the proportion of observations in each class the same in both subsets. Then one set is used for training and the other for testing, and then the sets are switched and retrained and retested. The datasets are briefly described in Table \ref{tab:ucisets}. The UC Irvine sets are chosen since they have been used in many machine learning studies \cite{ dietterich00, quinlan96} and are used by Banfield et al. \cite{banfield04} to compare bagging with boosting. The UC Irvine sets are taken from a wide variety of applications, so they also present a good breadth of data to test the versatility of methods.\\

\begin{table} \centering
\begin{tabular}{r l r r r}
 Set & Name & Attributes & Observations & Classes\\ \hline
 1 & breast-w & 9 & 699 & 2 \\ 
 2 & glass & 9 & 214 & 7 \\ 
 3 & ion & 34 & 351 & 2 \\ 
 4 & iris & 4 & 150 & 3\\ 
 5 & pendigits & 16 & 10992 & 10\\ 
 6 & phoneme & 5 & 5404 & 2 \\ 
 7 & pima & 8 & 768 & 2 \\ 
 8 & sonar & 60 & 208 & 2 \\ 
 9 & vehicle & 18 & 846 & 4 \\ 
 10 & waveform & 21 & 5000 & 3 \\ 
 \end{tabular}
 \caption{Description of UC Irvine datasets used to compare ensemble methods on multi-class problems.} \label{tab:ucisets}
 \end{table}
 
Experiments using the rule ensemble method were run using Matlab\texttrademark 7.10 on a MacBook Pro with a 2.66 GHz Intel Core i7 processor.


\section{Multiple Class Classification Results}\label{sec:multi-class}

The rule ensemble method is designed for binary classification problems, but many datasets contain multiple classes that one needs to identify. To be applicable to classification in general, we need to extend the rule ensemble to many class problems. Decision trees easily extend to multiple classes but the regression performed to assemble the rules in the rule ensemble prevent the rule ensemble from being extended to classification problems where the classes are not ordered. To identify multiple classes with the rule ensemble method we use the one-versus-all (OVA) classification technique that has been used for successfully extending many binary classification algorithms into multi-class algorithms \cite{hsu2002comparison, tan2003multi}. Other methods for extending binary classification algorithms to multiple class problems exist, such as all-versus-all classification. However, these methods require a large number of models to be built and are thus more expensive than OVA and frequently provide no more utility than the OVA classification method \cite{rifkin04}. \\

For a problem with $J$ classes, OVA classification performs $J$ binary tests, where the $j$th test checks if an observation is a member of the $j$th class or not the $j$th. Each observation gets a vector label prediction $\hat{\mathbf{y}} \in \mathbb{R}^J$, where each entry $\hat{y}_j$ is from the binary test classifying the $j$th class versus any other class. The prediction $\hat{\mathbf{y}}$ is a vector of -1's with a single positive entry. The index of the positive entry is the class that the observation is predicted to be from. \\

To extend the rule ensemble method we perform $J$ binary tests and each test returns a real valued prediction $F_j$. In the binary problem the label $\hat{y}_j$ is predicted to be the sign of the real value returned. However, in this setting it is possible that one of the binary models will misclassify the observation and result in $F_j$ being positive for more than one value of $j$. If we just took the sign of each $F_j$ then we would have a vector $\hat{\mathbf{y}}$ with multiple positive entries, indicating the observation was in multiple classes. In the event that $F_j$ is positive for more than one value of $j$, we take the prediction to be the class that has the most definitive prediction, i.e. the class $j^*$ where $F_{j^*}$ is greater than any other class label prediction. Choosing the largest label prediction is sensible, since the more confident the algorithm is that the observation is in a certain class, the closer to $1$ the label prediction will be. The closer to $0$ a class prediction is, the less certain the algorithm is of the observation's class affinity. \\

Here we compare the rule ensemble method, using {\sc Pathbuild}, with results from bagging and boosting tree ensemble methods. To compare we employ 10 datasets from the UC Irvine data repository \cite{uci} and the testing method parameters previously used to compare various ensemble methods \cite{banfield04}. Bagging uses 1000 trees, boosting uses 50 and both employ random forests for growing trees in five 2-fold cross validations. Tree ensemble labels can be estimated by a voting procedure, the prediction is the class that most of the trees predict the observation to be part of, and an averaging procedure, the label is the average of the the predictions made by all the trees. Results for both methods are presented. Minimal tuning was used to run the rule ensemble method on different datasets. \\


\subsection{Results Using OVA Classification on Vehicle Dataset}
\begin{figure} \centering
 \includegraphics[width=.5\textwidth]{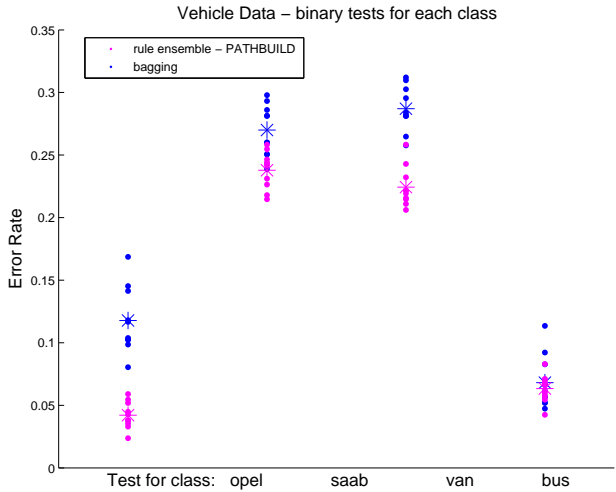} 
 \caption{Binary tests on each class of vehicle data. Figure shows accuracy when using bagging in an OVA classification method rather than with multi-class decision trees. } \label{fig:vehicle_results}
\end{figure}

Figure \ref{fig:vehicle_results} compares using the rule ensemble and bagging on the vehicle dataset. Bagging here is used in an OVA classification scheme rather than in its standard, direct multiple classification method. The error at predicting any given label in the set is shown. As can be seen in Figure \ref{fig:vehicle_results}, the rule ensemble beats bagging for the majority of the classes. Figure \ref{fig:vehicle_results} also shows the varying level of success that the ensemble techniques had at predicting each class. Some classes are easier to identify than others (e.g. ``opel" is easier to distinguish than van). Different ensembles were better suited to one class versus another, and which ensemble was better for a class was not consistent for all classes in a dataset.\\

\subsection{Results Using OVA Classification on All Datasets}
\begin{figure} [htb]\centering
	 \includegraphics[width=.5\textwidth]{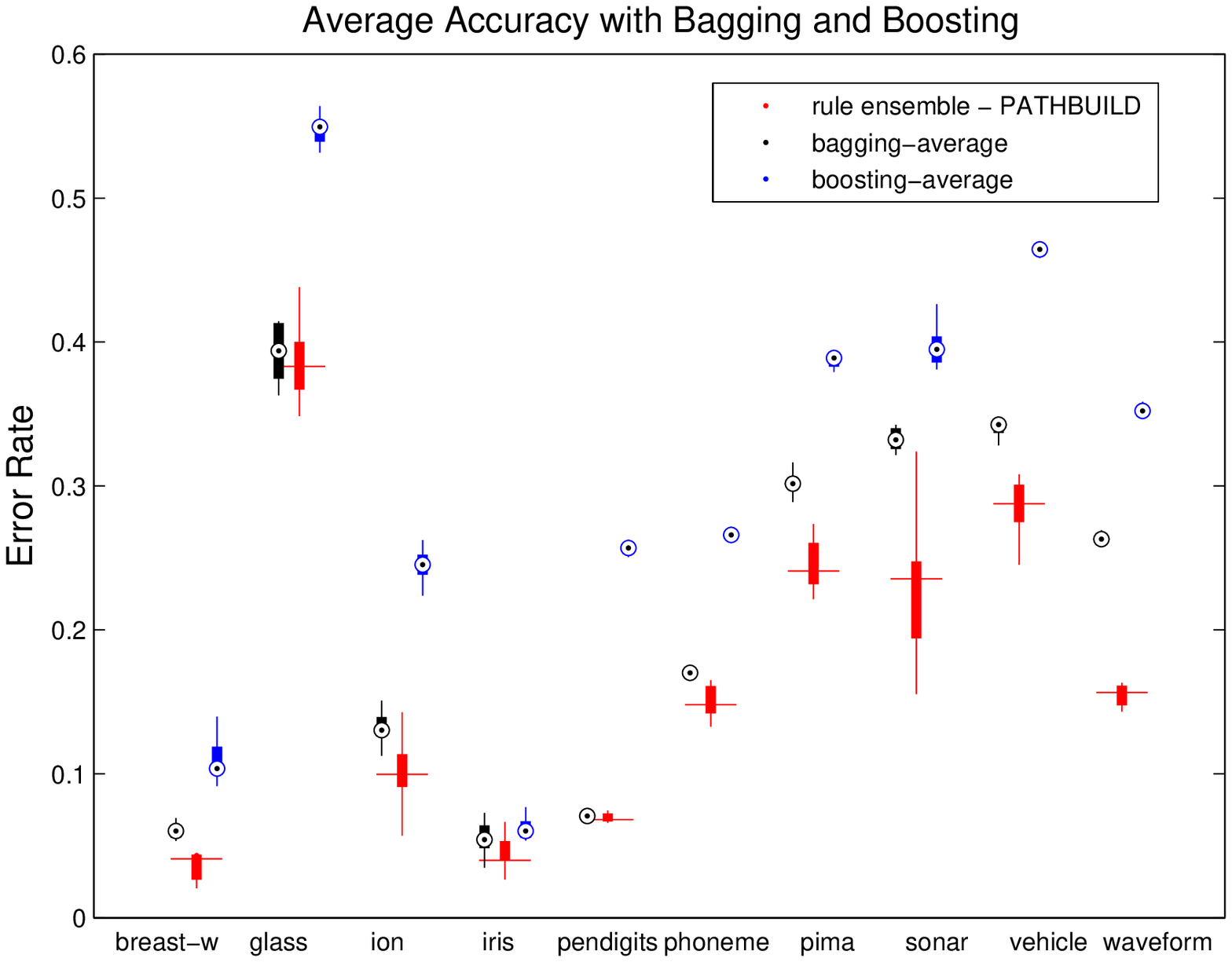}
	 \includegraphics[width=.5\textwidth]{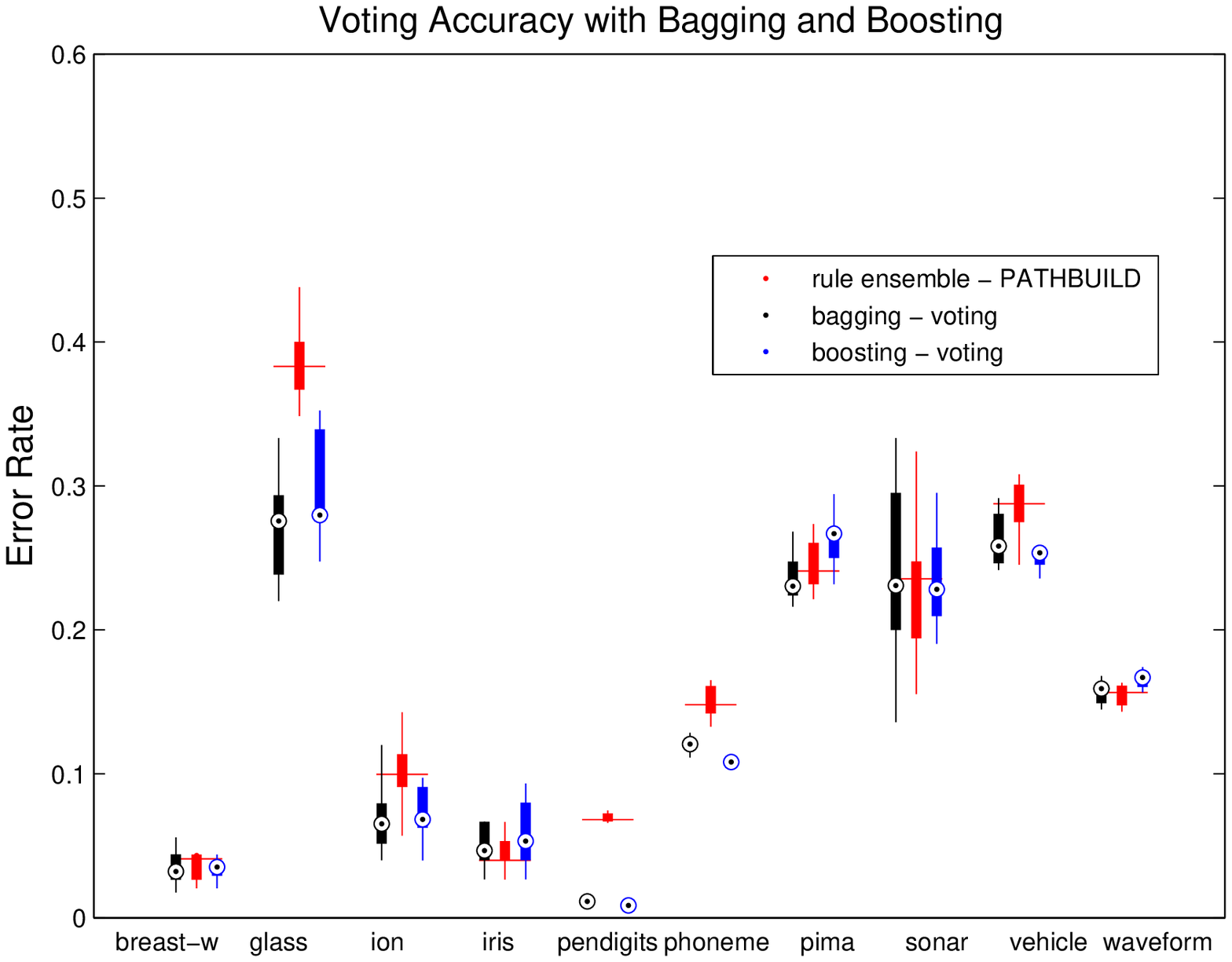}
	 \caption{Comparison of the error rate from a model that was generated with the rule ensemble method with the error rates from models that were generated with boosting and bagging ensemble methods. Results are summarized in Table \ref{tab:ucitable}.}
\label{fig:uci}	\end{figure}

\begin{figure}[htb] \centering
	 \includegraphics[width=.5\textwidth]{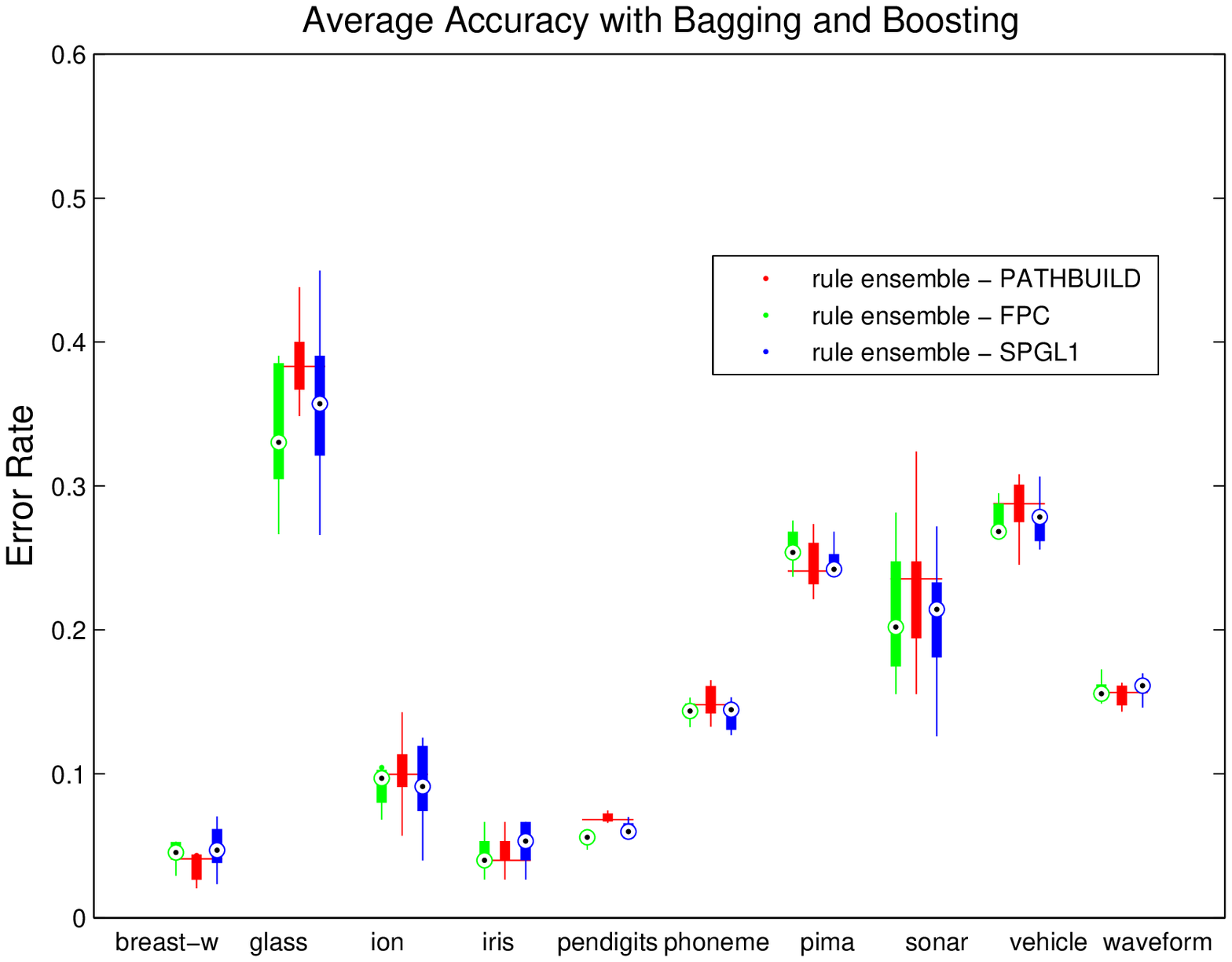}
	 \caption{Comparison of the error rate on 10 different datasets from models that were built with the rule ensemble method using different solvers are used to find the coefficients $\mathbf{a}$.}\label{fig:uci_solvers}	\end{figure}

The results of the multiple class tests are given in Table \ref{tab:ucitable}. The rule ensemble is much stronger than the tree ensembles if averaging of each tree's label prediction is used for classification. However, if the trees vote on which class label is best, then the rule ensemble is better on some datasets but not others. Voting clearly was better at label prediction than averaging base learner predictions, but neither boosting nor bagging provided a universal win over the rule ensemble, as can be seen in Figure \ref{fig:uci}. What is not apparent in Table \ref{tab:ucitable} is that the rule ensemble was a much better predictor for binary labels than the tree ensembles. This result is apparent in Figure \ref{fig:vehicle_results} where nearly every individual class is better predicted by the rule ensemble method. Figure \ref{fig:uci_solvers} shows the accuracy of the rule ensemble method with different coefficient solvers. Some datasets are easier to classify (larger percent of data correctly classified) while others, such as the \#2 dataset glass, were more difficult to classify for all the methods. \\

\begin{table}[ht] \centering
\begin{tabular}{l r || r r r | r r | r r}
& & \multicolumn{3}{c|}{Rule Ensemble} & \multicolumn{2}{c|}{Bagging} & \multicolumn{2}{c}{Boosting} \\ 
 Name & \# Classes &Pathbuild & {\sc Fpc} & {\sc SPGL1} & Voting & Average & Voting & Average \\ \hline
 
breast-w   &   2    &    4.34    &    4.60    &    4.75    &    3.51    &    5.97    &    \textbf{3.34}    &    10.93 \\ 
glass   &   7    &    37.99    &    33.47    &    35.26    &    \textbf{26.92}    &    39.16    &    29.83    &    54.72 \\ 
ion   &   2    &    9.97    &    10.43    &    9.23    &    \textbf{7.01}    &    13.44    &    \textbf{7.01}    &    24.49 \\ 
iris   &   3    &    4.80    &    \textbf{4.27}    &    5.33    &    4.93    &    5.51    &    5.73    &    5.95 \\ 
pendigits   &   10    &    6.94    &    5.65    &    6.10    &    1.23    &    7.05    &    \textbf{0.87}    &    25.68 \\ 
phoneme   &   2    &    14.97    &    14.33    &    14.16    &    12.06    &    16.97    &    \textbf{10.81}    &    26.61 \\ 
pima   &   2    &    24.45    &    25.76    &    24.56    &    \textbf{23.65}    &    30.13    &    26.22    &    38.78 \\ 
sonar   &   2    &    22.76    &    21.14    &    \textbf{20.67}    &    23.82    &    33.62    &    23.74    &    39.70 \\ 
vehicle   &   4    &    28.35    &    26.69    &    27.63    &    26.24    &    34.05    &    \textbf{25.18}    &    46.36 \\ 
waveform   &   3    &    \textbf{15.50}    &    15.79    &    16.03    &    15.67    &    26.30    &    16.61    &    35.26 \\  \hline
  \multicolumn{2}{l}{\textbf{Number of wins}}  &\textbf{1}&\textbf{1}& \textbf{1} & \textbf{3} & \textbf{0} & \textbf{5} & \textbf{0} \\
 \end{tabular}
 \caption{Error rate of the rule ensemble method compared with that of bagging and boosting. Error rate is given as the percent of observations in the test set that were misclassified.} \label{tab:ucitable}
 \end{table}

\section{Binary Classification Results}\label{sec:binary-class}

\subsection{Rule Ensemble with {\sc Pathbuild}}
Our implementation of the algorithm {\sc Pathbuild} for approximating the rule coefficients in the rule ensemble method is described in Figure \ref{fig:pathbuild pseudocode}. The coefficients are found by solving equation (\ref{eq:astar}) with a constrained gradient descent method. In this method, each iteration only advances in directions where the components of the gradient have magnitude greater than some fraction $\tau \in [0, 1]$ of the absolute value of the largest gradient component. Note that the set of directions we advance in,
$$ \{ k : |g_k(\mathbf{X}; \mathbf{a}^\ell)| \geq \tau* ||g_k(\mathbf{X}; \mathbf{a}^\ell)||_\infty\},$$ can change at every iteration. By not advancing in directions that have little change in the risk function, the expense of updating coefficients for variables of little importance is avoided. Not updating rules of little importance prevents the coefficient value for that rule from ``stepping" off zero, so that variable is effectively kept out of the model, allowing for a simpler model. Lower values of $\tau$ should include more rules in the model. The most inclusive model is when $ \tau = 0$, which is equivalent to using a basic gradient descent method to get a standard regression. Larger values of $\tau$ decrease the total number of rules used in the model. The most constrained model occurs when $\tau = 1$. 
 
\textbf{Effect of Number of Rules and Tree Size} \\

\begin{figure}
 \centering
 \includegraphics[width=.5\textwidth]{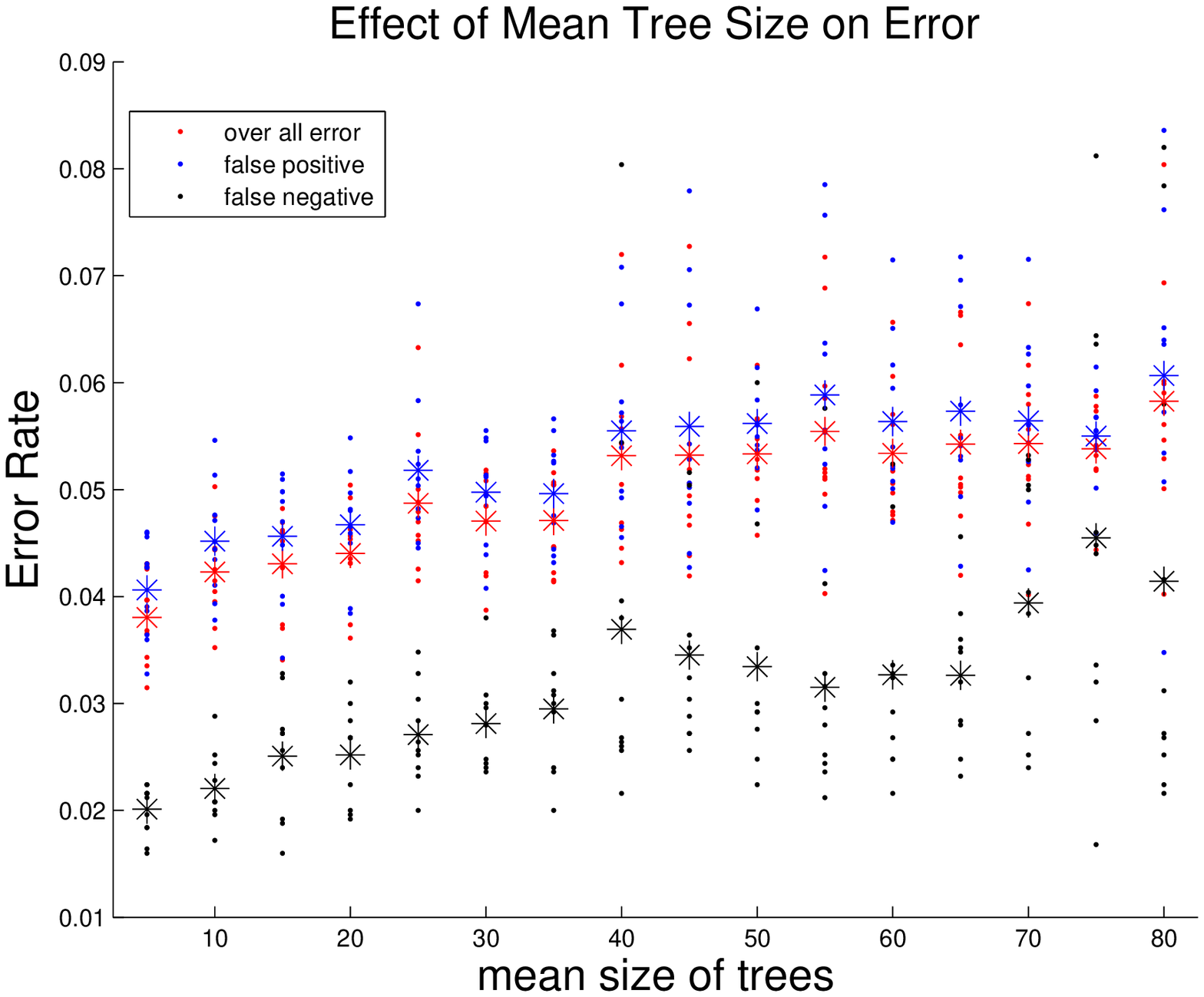} 
 \includegraphics[width=.5\textwidth]{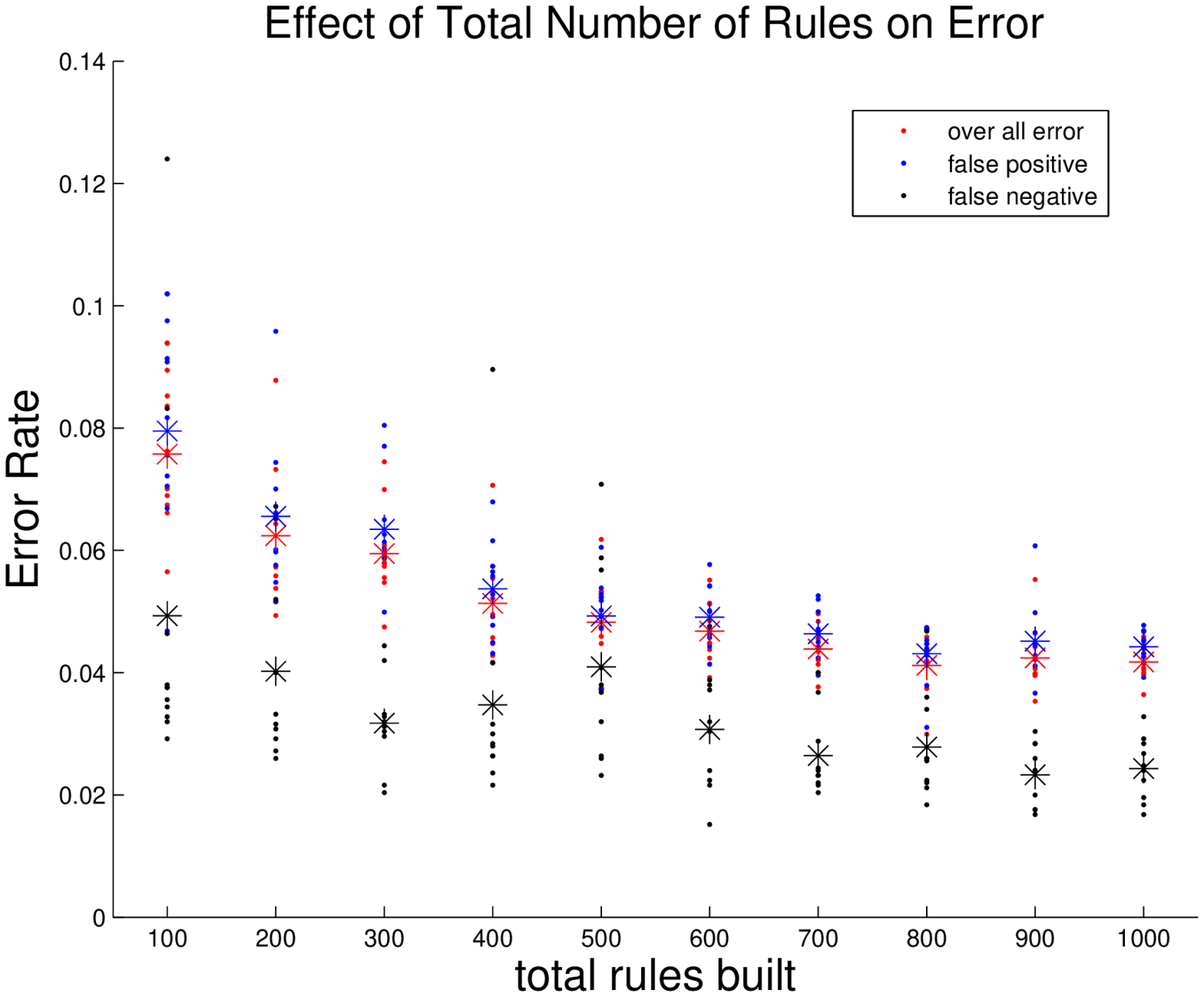} 
 \caption{The figure at the top shows that growing large trees (complex rules) increases the error rate. The bottom figure was made by growing trees with an average of 50 terminal nodes and shows that ensembles that have more rules have lower error rates. Tests were run with 500 maximum rules in each model. The $\tau$ tolerance was 0.5. Asterisks indicate the mean error rate from multiple tests.} \label{fig:REM_tuning}
\end{figure}

In Figure \ref{fig:REM_tuning} we see how the size of the trees and the number of rules used for the model affect the accuracy of the model. The decision trees are used to generate rules. Larger decision trees yield more complex rules than small trees because large trees have nodes that are deeper. Nodes deep in a tree capture subtle interactions within the training data since they depend on more splits and are more complex than nodes that are closer to the root node. Figure \ref{fig:REM_tuning} shows that ensembles built with larger trees have higher error rates than ensembles that use smaller trees. The increase in error rate when larger trees are built shows that when the model uses more complex rules, the model overfits the training data. However, the size of the trees does not have a strong effect on the how large of an error rate the rule ensemble has. Further, the accuracy of the rule ensemble is highly variable and the variance increases when larger trees are built. Ensembles built with trees that average 40 leaves had 4-7\% error, which is a large range when one considers that the mean classification error is only about 5.5\%. This error is larger than and has more variance than the error when trees with an average of 5 leaves are built, which is 3-4.2\% error. It is not clear why there is so much variance in the error rate in general. One should recall that the average number of terminal nodes in the decision trees are exponentially distributed, only the mean of the distribution is changed, so there is a variety of sizes of trees in each ensemble and complexity between rules in each ensemble. Because there is a variety of tree sizes there is some stability in the error rate as the mean size of the trees is changed. \\

The bottom of Figure \ref{fig:REM_tuning} also shows that using more rules can decrease the mean error rate of the rule ensemble method as well as the variance in the error rate. Increasing the number of rules built from 100 to 600 allowed the ensemble to capture more behavior and, as a result, nearly halved the error rate of the method. However, the error rate only decreases down to a certain point, after which adding more rules does not improve the model. For our data set, the error decreases to under 5.0\% when 600 rules are built, but does not continue to decrease substantially when more than 600 rules are used. We also see that the error rates between ensembles that are built on more rules have less variance than the error rates from ensembles that are built out of fewer rules. This result is reasonable, as having more rules gives the ensemble a better chance of finding good rules that successfully separate the data into classes. \\

In the initial tree building phase, a subsample of data is selected and a tree is grown on each random subsample. Our initial experiments took subsamples of 2,500 observations ($25\%$ of the total number of observations in the training set). When we decreased the subsample size to 500 observations ($10\%$ of training set), error rates did not significantly change even for a variety of tree sizes that had between 5 and 80 terminal nodes. The lack of significant difference indicates that 500 observations give us a large enough sample to catch the same amount of behavior that is captured when larger subsamples of data are used to build each tree. \\

\begin{figure}[h] \centering 
 \includegraphics[width=.5\textwidth]{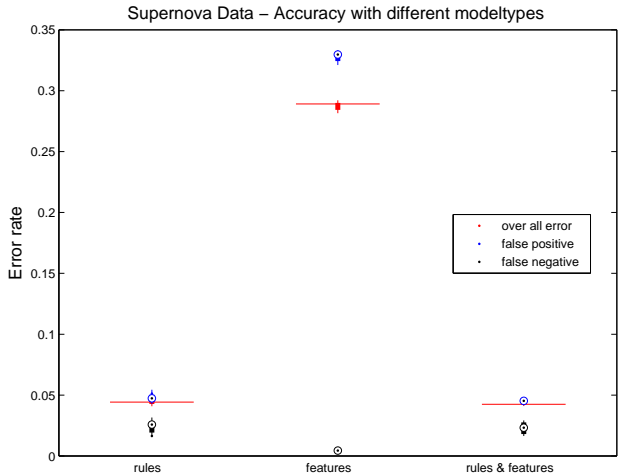} 
 \caption{ Using rules in an ensemble was six times more accurate than only using linear terms, a classical multiple linear regression of the labels on the attribute variables. Linear regression was not reliable for predicting the labels, but using the rule ensemble allowed for only 5\% error in prediction. This experiment was run using {\sc Pathbuild} to solve for coefficients.} \label{fig:modeltypes}
\end{figure}

\textbf{Effect of Using Rules Versus Linear Terms} \\
In Figure \ref{fig:modeltypes} we see the effect of allowing the model to have linear dependencies on individual features. If only linear terms are used, then the model is a standard multiple linear regression. Allowing the model to be built with both linear terms and the rules generated by the trees yields a mixed model. Using rules for the regression terms provides a clear advantage over the standard regression model by reducing the error rate from nearly 30\% error to less than 5\%. The linear regression is also more biased in its error than the rule ensemble. This bias can be seen by the false negative rate being close to zero; this means nearly all the error is caused by mislabeling observations with negative labels. We would not expect a linear regression to capture any of the complex nonlinear behavior in the dataset, and the error rates show that such an conjecture is correct -- rules are needed to get significant predictive capability.\\

\begin{figure} \centering
\includegraphics[width =.5\textwidth]{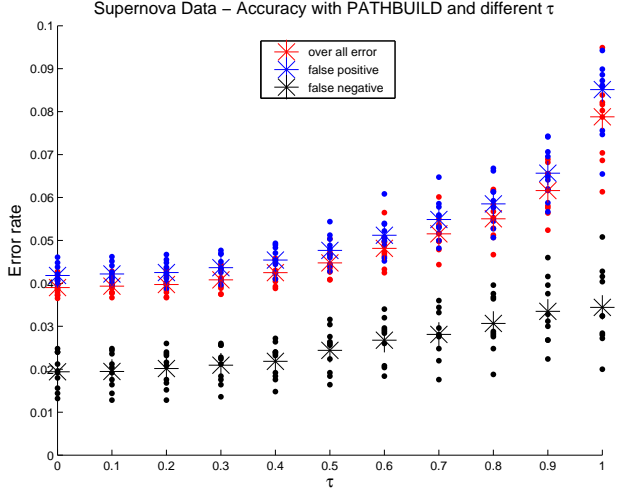}
\caption{Error rate increases as $\tau$ increases and restricts the number of coordinates the algorithm advances in at each iteration. This experiment was run with each tree having an average of 20 terminal nodes and 600 maximum rules.} \label{fig:tau_error_rates}
\end{figure}

\textbf{Effect of Using the $\tau$ Threshold as Penalty}\\
The variable $\tau$ controls how many directions are updated at each iteration of {\sc Pathbuild} in the thresholded gradient descent method. The results of increasing $\tau$ are shown in Figure \ref{fig:tau_error_rates}. The model becomes less accurate and the variance of the error rate increases, as $\tau$ increases. An increase in $\tau$ causes a higher threshold that results in fewer terms being included in each iteration of the coefficient finding method and a ensemble model that is less accurate. It is interesting to note that within a certain range, decreasing $\tau$ further does not offer much increase in the predictive capability of the model. In this example, we see that when $\tau$ is between 0 and 0.3 there isn't a large increase in error rate. This indicates that using a weaker threshold of $\tau = 0.3$ or even $\tau = 0.4$ will not significantly compromise the accuracy of our model. This is a good result, as using a larger threshold decreases the computational expense of each iteration of the gradient descent method. The result that $\tau = 0.3$ produces similar error rates to using $\tau = 0$ means that we can get the same accuracy with less computation.

\subsection{Rule Ensemble with {\sc Glmnet}}\label{sec:glmnet}
In this experiment we use the {\sc Glmnet} package \cite{friedman09}, which returns approximations to solutions of elastic-net regularized general linear models, to solve for the coefficients $\mathbf{a}$ within the rule ensemble method. {\sc Glmnet} approximates a solution to the least squared error regression subject to an elastic net penalty, which is 
\begin{equation} \label{eq:elasticnet}
\underset{ \mathbf{a} \in\mathbb{R}^n} \min \frac{1}{N} ||\mathbf{X}\mathbf{a}-\mathbf{y}||_2 + \lambda P_{\alpha}(\mathbf{a}),
\end{equation}
with a coordinate-wise gradient descent method \cite{friedman09}. The elastic net is defined as
$$P_{\alpha}(x) = \alpha ||\mathbf{a}||_1 + (1-\alpha)||\mathbf{a}||_2^2$$ 
for $\alpha \in [0,1]$. When $\alpha = 0$ the problem is referred to as ridge regression, and when we set $\alpha = 1$ we get the same problem as in equation (\ref{eq:regularizedreg}). The coordinate-wise gradient descent method starts with the null solution, similar to {\sc Pathbuild}, then cycles over the coefficients and uses partial residuals and a soft-thresholding operator to update each coefficient one by one \cite{friedman07}. {\sc Glmnet} has some modifications that also allow some parameters to be associated with and updated at the same time as neighboring parameters. The null solution corresponds to solving equation (\ref{eq:elasticnet}) with $\lambda = \infty$. As the coefficients are updated, $\lambda$ is decreased exponentially until the lower bound $\lambda_{min}$, the desired and pre-specified penalty weight, is met. {\sc Glmnet} calculates a set of coefficients along each increment of the path $\lambda = \infty$ to $\lambda = \lambda_{min}$ and uses the previous solution as a ``warm start" to approximate the next solution. Note that $\lambda_{min}$ should be small enough to prevent the penalty from being so large that it causes the vector to be overly sparse. However, $\lambda_{min}$ should also be positive and large enough to ensure a sparse solution that is robust to the training data. A robust solution includes terms for interactions that are inherent to the application generating the data, not interactions that are only figments the subset selected for training. It is not clear how to pick the penalty weight $\lambda$ to maintain sparsity of the solution and prevent overfitting while also capturing enough characteristics of the dataset. \\ 

\begin{figure} \centering
	\includegraphics[width =.5\textwidth]{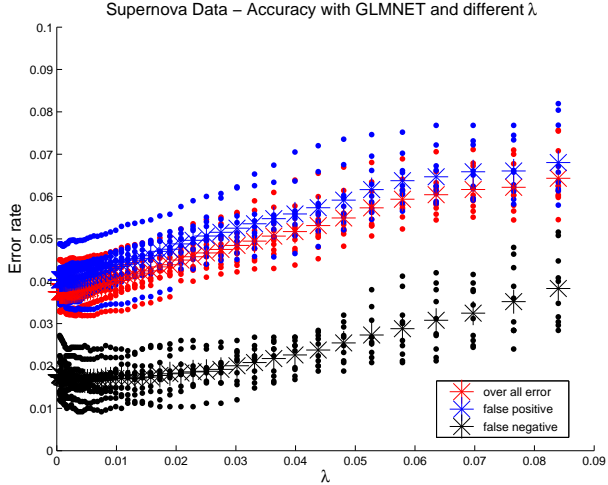}
\caption{Error rate decreases when {\sc Glmnet} is used to solve for the coefficients and the constraint parameter $\lambda$ decreases.}\label{fig:glmnet_error_rates}
\end{figure}
Here we use the rules generated in the previous experiment with {\sc Glmnet} and build models using the coefficients that are generated at each step of the path $\lambda \in [\lambda_{min}, \infty]$. Figure \ref{fig:glmnet_error_rates} shows how the accuracy of the method changes as the weight of the penalty used to find the coefficients changes. The solution with {\sc Glmnet} when $\lambda$ is small results in slightly less error than the solution with {\sc Pathbuild} when $\tau$ is small. The variance in the error rates from solutions found with {\sc Pathbuild} is less than the variance of error rates from solutions found with {\sc Glmnet}. Both solutions yield false positive rates that are more than twice as large as the false negative rates; this is probably a result of the ratio of positive to negative observations in the test set is small. The error rate slowly decreases as $\lambda$ decreases, but then the error rate stabilizes when $\lambda$ is very small, $ < 0.01$. It is interesting that the variance in error rates of the solutions is relatively constant as $\lambda$ changes.\\

\subsection{Rule Ensemble with {\sc Spgl1}}\label{sec:spgl1}
In this experiment, we used the {\sc Spgl1} (sparse projected-gradient $l$1) Matlab\texttrademark package \cite{spgl1} to solve for the coefficients $\mathbf{a}$ in 
\begin{equation}
\min || \mathbf{X} \mathbf{a}-\mathbf{y}||_2 \text{ subject to } || \mathbf{a} ||_1 <= \sigma.
\label{eq:spgl1}\end{equation}
At each iteration of the algorithm, a convex optimization problem is constructed, whose solution yields derivative information that can be used by a Newton-based root-finding algorithm \cite{berg08}. Each iteration of the {\sc Spgl1} method has an outer/inner iteration structure, where each outer iteration first computes an approximation to $\sigma$. The inner iteration then uses a spectral gradient-projection method to approximately minimize a least-squares problem with an explicit one-norm constraint specified by $\sigma$. Some advantages of the {\sc Spgl1} method are that only matrix-vector operations are required and numerical experience has shown that it scales well to large problems.

\begin{figure}  \centering
  \includegraphics[width =.5\textwidth]{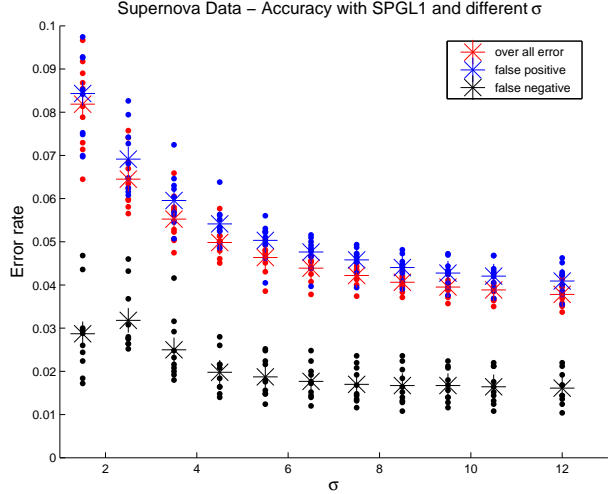} 
  \caption{Error rate decreases when {\sc Spgl1} is used to solve for the coefficients and the constraint parameter $\sigma$ increases.}  \label{fig:spgl1_error_rates}
\end{figure}

The results using {\sc Spgl1} are shown in Figure \ref{fig:spgl1_error_rates}. The accuracy of the {\sc Spgl1} solution increases when $\sigma$ increases. The error rates are similar to those found by {\sc Pathbuild} and {\sc Glmnet}, but slightly higher than {\sc Glmnet} even when $\sigma$ is large. 

\subsection{Rule Ensemble with {\sc Fpc}}\label{sec:fpc}
In this experiment, we used a fixed point continuation method ({\sc Fpc}) \cite{hale07} that approximates the solution $\mathbf{a}$ in 
\begin{equation}
\underset{ \mathbf{a} \in\mathbb{R}^n} \min \text{ } ||\mathbf{a}||_1 + \frac{\mu}{2}*||\mathbf{Xa} - \mathbf{y}||_2^2. 
\label{eq:Fpc} \end{equation}
This problem formulation seeks to minimize the weighted sum of the norm of the coefficients and the error of the solution, the left and right terms respectively. The sparsity of $\mathbf{a}$ is controlled by the size of the weighting parameter $\mu$. Increasing $\mu$ places more importance on minimizing the error, and reduces the ratio of the penalty to the error. The reduction of penalty importance allows more coefficients to become non-zero (the $\ell_1$ norm of the coefficients to increase) and thus find a closer fit to the problem. Equation (\ref{eq:Fpc}) is simply a reformulation of problem (\ref{eq:regularizedreg}) with the lasso penalty, and is referred to as a basis pursuit problem in signal processing. The relation of the two problems can clearly be seen if, for any $\lambda$ value, $\mu$ is chosen to be
$$\mu = \frac{2}{N \lambda}$$ 
and equation (\ref{eq:Fpc}) is multiplied by $\lambda$. {\sc Fpc} was developed for compressing signals by extracting the central components of the signal.\\

{\sc Fpc} exploits the properties of the $l_2$ norm and declares three equivalent conditions for reaching an optimal solution. {\sc Fpc} uses the reformulations of the optimality conditions to declare a shrinkage operator $s_{\nu}$, where $\nu $ is a shrinkage parameter that has both an effect on the speed of convergence and how many non-zero entries $\mathbf{a}^*$ has. The operator $s_{\nu}$ acts on a supplied initial value $\mathbf{a}^0$ (which we chose to be the null solution) and finds our solution $\mathbf{a}^*$ through a fixed point iteration
$$\mathbf{a}^* = s_{\nu}(\mathbf{a}^*).$$
The given condition for the threshold of $s_{\nu}$ is $$ \text{if } \nu-|\mathbf{y}| > 0 \text{ then } s_{\nu}(\mathbf{y}) \rightarrow 0.$$
{\sc Fpc} forms a path of solutions that starts with $\mu$ initialized to $ \mu = \frac{ \eta } { ||X^t \mathbf{y}||_ {\infty}} $ (where $\eta$ is a ratio of possible optimal square error at the next step to the square error at the current step). The parameter $\mu$ is altered at each step, which forces the shrinkage parameter to expand and contract but the upper bound for $\mu$ is supplied by the user. All results presented here use {\sc Fpc} with projected gradient steps and optionally using a variant of Barzilai-Borwein steps \cite{barzilai88}.\\

\begin{figure} \centering
	 \includegraphics[width=.5\textwidth]{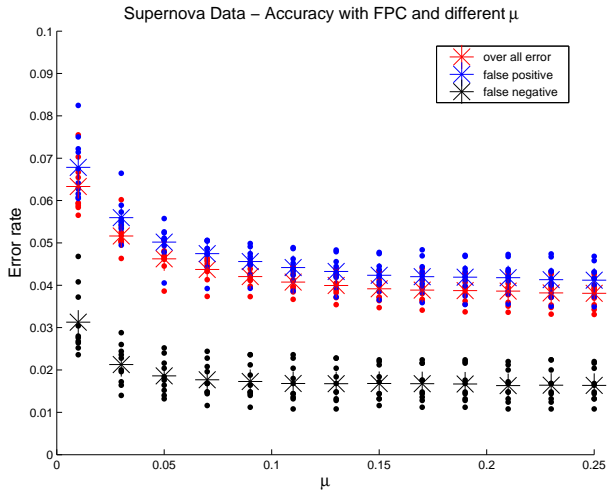}
\caption{Error rate decreases when {\sc Fpc} is used and the weight on the risk or mean squared error is increased ($\mu$ increased).}\label{fig:fpc_error_rates}
\end{figure}

\begin{figure}  \centering
  \includegraphics[width =.5\textwidth]{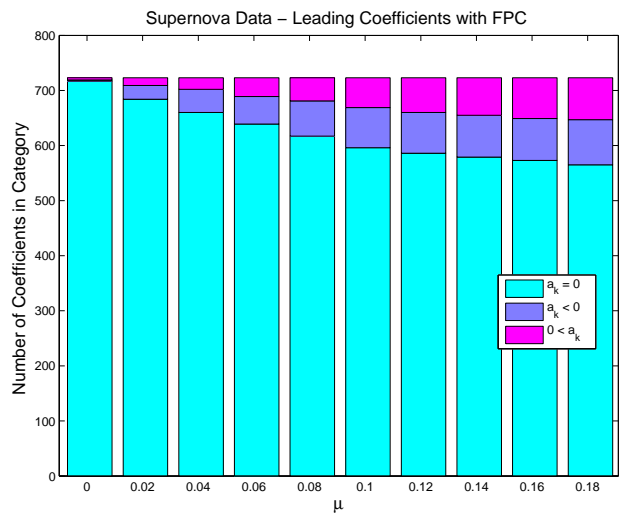}
 \caption{Sparsity of solution is indicated by the number of coefficients not equal to zero. As $\mu$ is increased, the solution becomes less penalized and more coefficients step off zero and allow more terms to be included in the model. The sparsity of the solution stops decreasing when $\mu$ is large and the penalty is relatively small compared to the emphasis on minimizing the risk or second term in equation (\ref{eq:Fpc}). Here 78\% of the coefficients are trivial when $\mu = 0.19$.}  \label{fig:fpc_coeff_by_category}
\end{figure}

The results of solutions generated by {\sc Fpc} are shown in Figure \ref{fig:fpc_error_rates}. They are roughly as accurate as the solutions generated with the previous solvers. {\sc Fpc} also has an explicit display of the thresholding as seen in Figure \ref{fig:fpc_coeff_by_category}; the norm of the coefficients increases dramatically then asymptotically approaches a certain value. The asymptotic behavior is caused by the threshold constricting the coefficients and essentially preventing another coefficient from stepping off of zero. The thresholding is also seen in the error rate decreases as the weight on the mean squared error is increased, but stabilizes once the training set is reasonably fit. The value of $\mu$ where the error stabilizes is the value needed to build the model, but unfortunately it is not clear how to choose this value of $\mu$  {\em a priori}. The need for a selection of the penalty parameter is one of the difficulties that {\sc Fpc}, { \sc Spgl1}, and {\sc Glmnet} have. {\sc Pathbuild} shares a similar problem with the need to selection the gradient descent constriction parameter $\tau$.

\begin{figure}  \centering
  \includegraphics[width =.5\textwidth]{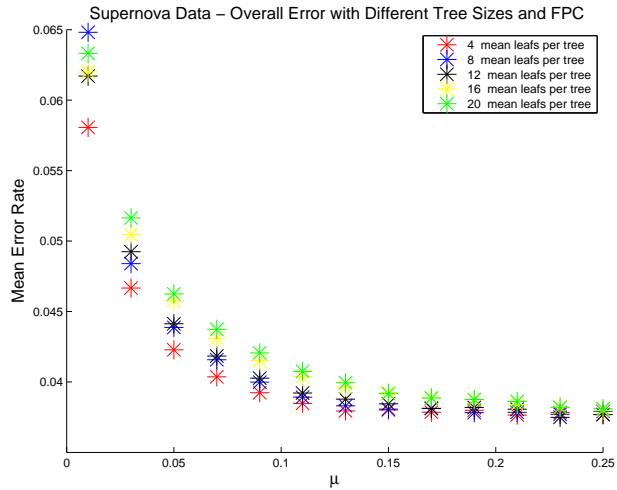} 
  \caption{There is little fluctuation in the overall error rate when {\sc Fpc} is used on rules that were built with different size trees. Only the mean of cross validation tests is plotted here for simplicity. Little fluctuation implies that simpler rules, which come from smaller trees, can be used to build a model without sacrificing predictive capability.}  \label{fig:fpc_error_rate_change_treesize}
\end{figure}

\subsection{Identifying Important Attributes Via Rule Importance}
Figure \ref{fig:fpc_error_rates} shows that the rule ensemble method is quite successful at correctly classifying observations when all of the attributes are used to generate rules and build the model. Attributes have variable importance in the model and we suspect that not all of the 39 attributes in the full dataset are needed to model and correctly predict class labels. We want to use the rule ensemble method to select only the attributes that are important and save the expense of considering the other less important variables.\\

\begin{figure} \centering
	 \includegraphics[width=.5\textwidth]{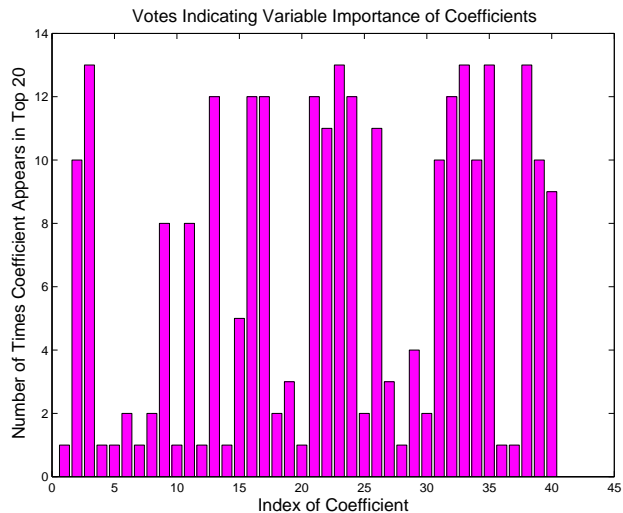}
\caption{This histogram shows how many times a rule was one of the top 20 most important rules in a solution. A solution was generated at each of 13 different values of $\mu$, as shown in Figure \ref{fig:fpc_error_rates}. Rules that received 13 votes were one of 20 most influential rules for every value of $\mu$ tried. Only rules that were in the top 20 most influential for at least one solution are shown. The attributes that were used in the rules shown here were used to find a smaller subset of attributes to train on for the results in Figure \ref{fig:compare_error_rate_retest}.}\label{fig: fpc_hist_votes_for_coeff}
\end{figure}

The importance of a rule is indicated by the magnitude of the coefficient for that rule. The larger a coefficient is in magnitude, the more important the corresponding rule is, since that rule will have a larger contribution to the model. To sift out the most important attributes, we look at which rules {\sc Fpc} considered important at different values of $\mu$. Rules are ordered by the magnitude of their corresponding coefficient and the rules corresponding to the 20 largest (in magnitude) coefficients are selected. An example of ordering the rules is in Table \ref{tab:table_of_rules} where the 5 most important rules from one test are ordered. This process is continued for 5 different repetitions of training and testing, which yields 5 sets of 20 most important rules. The sets of rules are decomposed into sets of attributes that are used to make up the rules in each set. Then we let the 5 repetitions vote on which attributes are influential and keep only attributes that are in the set of important attributes for at least 3 out of the 5 repetitions. Figure \ref{fig: fpc_hist_votes_for_coeff} shows how many votes the highest ranking rules get and indicates that certain rules are important in all solutions while others are considered important in only some solutions. This set of attributes forms a smaller subset of the 39 attributes available in the initial dataset. The subset of rules only contains attributes that are used in at least one of the 20 most important rules in at least 3 of the 5 repetitions. \\

The importance of a rule is indicated by the magnitude of the coefficient for that rule. The larger a coefficient is in magnitude, the more important the corresponding rule is, as that rule will have a larger contribution to the model. To sift out the most important attributes, we look at which rules {\sc Fpc} considered important at different values of $\mu$. Rules are ordered by the magnitude of their corresponding coefficient and if a rule is one of the top 20 most important in a solution generated with a certain $\mu$ (13 values of $\mu$ we considered), then that rule receives a vote. An example of ordering the rules is in Table \ref{tab:table_of_rules} where the 5 most important rules from one test with a given $\mu$ are ordered. Figure \ref{fig: fpc_hist_votes_for_coeff} shows for how many values of $\mu$ each rule was considered to be in the top 20 most important; this indicates that certain rules are important in solutions with all values of $\mu$ tried while others are considered important only when certain $\mu$ are used. This process is continued for 5 different cross-validation sets, which yields 5 sets of rules that were in the top 20 most important rules for at least one value of $\mu$. The sets of rules are decomposed into sets of the attributes that were used to make up the rules in each set. Then we let the 5 repetitions vote on which attributes are needed to make the most influential rules and keep only the attributes that are in the set of important attributes for at least 3 out of the 5 repetitions. This set of attributes forms a smaller subset of the total attributes available in the initial dataset; it is the subset attributes that are used in at least one of the most important rules in at least 3 of the 5 repetitions. \\

\begin{figure} \centering
	 \includegraphics[width=.5\textwidth]{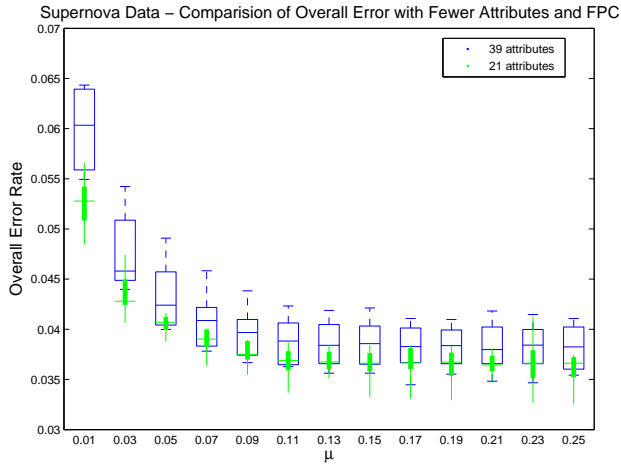}
\caption{Comparison of overall error rate when fewer attributes are used. The preliminary tests used all 39 attributes in the dataset. The subsequent tests used only the subset of 21 attributes that were used to construct the most important rules in the preliminary tests. Using the restricted set of attributes gives a lower error rate indicating that the rule ensemble method successfully identified which attributes are important in the dataset. }\label{fig:compare_error_rate_retest}
\end{figure}

For the supernova dataset, the smaller subset of attributes included only 21 of the 39 original attributes. Tests were repeated using the same cross-validation sets and method parameters as were used in Figure \ref{fig:fpc_error_rates}, but using only the smaller subset of 21 attributes to train on rather than all 39 attributes. Figure \ref{fig:compare_error_rate_retest} compares the error rate of the method when 21 attributes were used with the error rate of the method when all 39 attributes were used. The results show that the accuracy of the method improves when we reduce the number of attributes used in the model. The method successfully ranks rules and identifies more important attributes. The method loses accuracy when the less important features are included; in essence, the extra attributes act as noise. After the method identifies these attributes as less important and we remove them, the method is able to return an even more accurate model and the insight of which attributes are not adding predictive capability to the model. Garnering better accuracy with fewer attributes may allow the extra attributes to be excluded from the data collection, which will save time in collecting data, save space in storing data, and allow an overall better analysis.\\

\begin{table} \centering
\begin{tabular}{c r}
Rule $r_k$             & $|a_k|$ \\ \hline
$x_2 \geq $ -0.315 \& $x_{18} \geq $ 0.047 & 0.1045\\
$x_{29} < $-0.251           		& 0.0725\\
$x_{23} \geq$ -0.606         		& 0.0317\\
$x_1 < $ -0.324           		& 0.0274\\
$x_{12} \geq$ 0.260          		& 0.0193\\
\end{tabular}
 \caption{Example of ordering rules by importance. These are the five rules with greatest importance in the first model as selected by {\sc Fpc} with $\mu = 0.25$. } \label{tab:table_of_rules}
 \end{table}

\section{Conclusions}
 \begin{figure}[htb] \centering
	 \includegraphics[width=.5\textwidth]{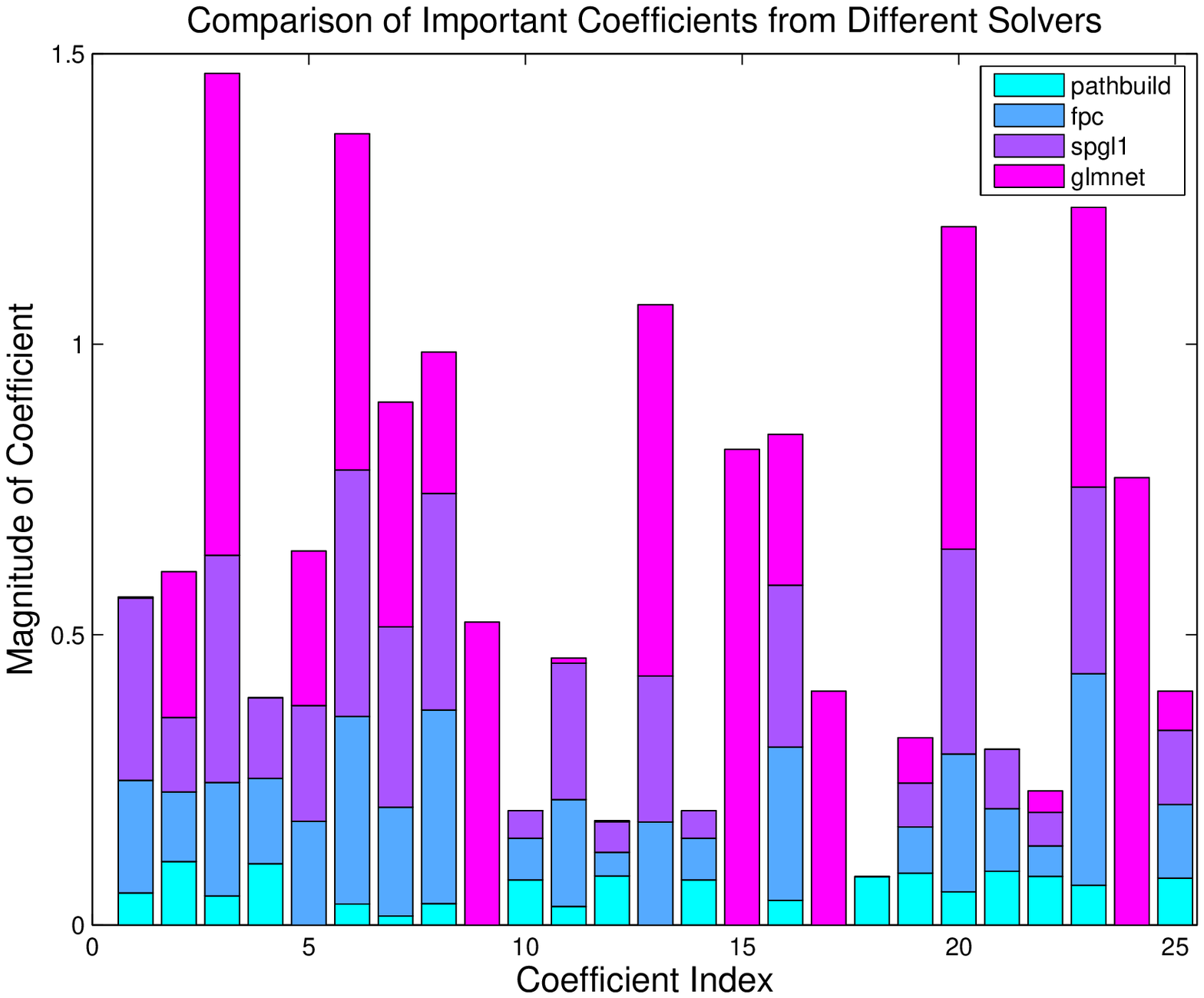}
\caption{Length of bars indicate the magnitude of coefficients as calculated by different solvers. Only the coefficients with the 10 largest magnitudes from each solver are displayed. Coefficients plotted come from solutions that yielded similar error rates: $\tau = 0.4$ , $\mu = .11$, $\sigma = 8.5$, $ \lambda = 0.014$.}\label{fig:compare_important_coeff_3solvers}
\end{figure}

We compared several variations of a rule ensemble method with some well-known tree ensemble methods, namely boosting and bagging, on a variety of multi-class problems. We extended the rule ensemble to work on multi-class problem by using the OVA technique and found that with this extension the rule ensemble method performed comparably to theÊtree methods on a set of 10Êclassical datasets. This result highlights the power of the rule ensemble method, as we had expected the tree ensemble methods to do better on multi-class problems. Tree ensembles can use multi-class decision trees, which provide what one would think is a more natural extension to multi-class problems than using the OVA method. However, the rule ensemble method returned comparable rates of accuracy on most datasets and even performed better on some of the datasets. The discrepancy between the tree ensembles with voting and the rule ensemble was larger on problems that had a relatively large number of labels, such as the pendigits dataset, which had the most labels out of all the datasets, than on datasets with fewer labels. To improve the accuracy of the rule ensemble on problems with many classes, we would like to try using multi-class decision trees to build the rules and then relabel the nodes for each binary problem. This technique might yield better rules as it would allow for differentiation between the classes in the rule building phase. Better rules would then allow for a clearer separation of binary labels in the regression phase. This technique would also make the training phase more efficient as it would only require one set of rules to be constructed rather the as many sets of rules as there are classes.\\

We also looked at using 4 different methods to find coefficients to assemble the rules. All 4 methods present the challenge of needing to select a constraint parameter that controls the sparsity/accuracy trade-off of the solution that they return. If each parameter is chosen correctly then the methods are capable of producing coefficients that allow for similar accuracy in the model. The different approaches that the methods take for finding the coefficients do result in slightly different rankings of the rules. The difference in coefficients that each method considers important is shown in Figure \ref{fig:compare_important_coeff_3solvers}. Ideally all solvers would select the same terms to be the most significant and would order the terms by importance the same way. Figure \ref{fig:compare_important_coeff_3solvers} shows that some rules that one method considers important are not considered to be important to another method. {\sc Fpc} and {\sc Spgl1} order coefficients similarly, which is indicated by {\sc Spgl1} giving a significant magnitude to coefficients that {\sc Fpc} also gives a significant magnitude to. {\sc Glmnet}'s and {\sc Pathbuild}'s ordering share less similarity with {\sc Fpc} and {\sc Spgl1} as indicated by coefficients such as 9 and 18 that {\sc Glmnet} and {\sc Pathbuild} give a significant magnitude to, but both {\sc Fpc} and {\sc Spgl1} give trivial values to. The difference in methods is also reflected in the sparsity of the solutions that they return. To achieve similar accuracy (taken here at 96\% accuracy) {\sc Pathbuild} returns a solution with 40-50\% of the coefficients non-zero while the other methods return much sparser solutions that have only 12-19\% of the coefficients non-zero. In general, {\sc Spgl1} returned the sparsest solutions and {\sc Pathbuild} returned the least sparse solutions for models with similar error rates.\\

As a final step, we showed the utility of the rule ensemble method for identifying important attributes in a dataset containing images of potential supernovas. The rule ensemble method has the benefit over tree methods of providing insight into a dataset by returning weighted rules. Rules with large weights have a larger effect on the model and thus can be thought of as more important than other rules. We used the importance of such rules to alert us to the more significant features in the dataset by looking at which features the important rules are defined on. This technique allowed us to select 21 attributes out of the 39 available and reduce the error rate of the model by building models only on the reduced set of attributes. Traditional algorithms that use ensembles of decision trees, such as boosting and bagging, aren't able to provide this insight into the importance of certain variables of a dataset because they do not rank or weight of rules. \\

The rule ensemble method has the advantage over some other methods by being able to identify relationships and hierarchies between variables to a certain extent when building the decision trees. The rules in the decision trees get more complex the deeper the tree is grown and also are able to have limited support in the parameter space, so they only affect certain observations that fall in that space. By including more variables, complex rules can be seen as resembling discrete correlations, and the post-processing of the rules allows for overly simplified correlations (that precede more complex rules in depth) to be removed from the model. The post-processing also allows for overly complex rules to be pruned from the model. Thus some variable interactions can be captured by the rule ensemble method without any {\em a priori} assumption that they exist, as is needed in standard regression models, and excessive computation is not spent considering correlations that do not exist.\\

We do not compare the computational efficiency of the rule ensemble method with tree ensembleÊmethods here, since it is currently written in Matlab\texttrademark, while the tree ensemble methods used are written in C. However, we do not expect that the rule ensemble method will reduce the amount of time necessary for the training portion of the algorithm to run because it must perform the coefficient solving method in addition to the tree growing. If the rule ensemble method is able to prune a substantial number of repetitive or unnecessary rules, thenÊit is likely to run substantially more quickly than the tree methods. Comparing the time efficiency of the rule ensemble with other tree methods and other machine learning techniques will be part of future work. We do not present the computationalÊefficiency of the coefficient solving methods used inÊthe rule ensemble method for the same reason. ÊEach solver is written in aÊdifferent programming language, and each will have to be implemented in the same language and level of optimization before a meaningful study can be performed.

\section*{Acknowledgements} 
We would like to thank Sean Peisert and Peter Nugent for their valuable comments and suggestions.

\clearpage

\appendix
\newcommand{\appsection}[1]{\let\oldthesection\thesection
  \renewcommand{\thesection}{Appendix \oldthesection}
  \section{#1}\let\thesection\oldthesection}

Here we discuss the gradient method Pathbuild, which is described in section \ref{sec:Pathbuild}, in greater detail. Simplifications of the gradient method are presented and considered as the ``fast method".
\section{Derivation of the Negative Gradient of Risk}
\label{loss_derive}
The negative gradient $\mathbf{g} \in \mathbb{R}^{K}$ of the loss on the observations is found by taking partial derivatives of the sum of the loss on each observation with respect to each coefficient. The components of the negative gradient are given by
\begin{equation} \label{gradient}
g_k=-\frac{\partial}{\partial{a_k}}\frac{1}{N}\sum_{i=1}^NL(y_i,F(\mathbf{x}_i)),
\end{equation}
	where $k = 1,\dots, K$. Note that $g_0 = 0$ as $a_0$ is the constant intercept that minimizes the risk when $F(\mathbf{x}_i) = a_0$ and all the other coefficients have not moved off their initial zero value. $\{g_k, \, k = 1..K\}$ are the non trivial components of the gradients.
\begin{equation} \label{DLDa}
\frac{\partial{}}{\partial{}a_k}L(y_i,F(\mathbf{x}_i))=\frac{\partial{}L(y_i,F(\mathbf{x}_i))}{\partial{}F(\mathbf{x}_i)}\frac{\partial{}F(\mathbf{x}_i)}{\partial{}a_k}.
\end{equation}
Note that the second term is easily computed from the linear form of $F(\mathbf{x}_i)$ and is given by 
\begin{equation} \label{DFDa}
\frac{\partial{}F(\mathbf{x}_i)}{\partial{}a_k}=x_{ik}.
\end{equation}


\subsection{Negative gradient squared error ramp loss is used}
The previous discussion has been generalized for the use of any loss function $L(\cdot)$. Now consider the case when the loss function is given by
\begin{equation} \label{ramp}
\begin{array}{c}
L(y_i,F(\mathbf{x}_i))=[y_i-H(F(\mathbf{x}_i))]^2\\ 
H(F(\mathbf{x}_i))=\max[-1,\min(1,F(\mathbf{x}_i))],
\end{array}
\end{equation}
which is the squared error ramp loss for the $i$-th observation. We want to find the derivative with respect to $\mathbf{a}$ for this loss function. Begin by taking a partial derivative with respect to $F$
\begin{equation} \label{DrampDF}
\frac{\partial{}}{\partial{}F}L(y_i,F(\mathbf{x}_i))=-2(y_i-F(\mathbf{x}_i))\,I(|F(\mathbf{x}_i)|<1).
\end{equation}
Substitute (\ref{DFDa}) and ({DrampDF}) into (\ref{DLDa}) to get the derivative for the squared error ramp loss 
\begin{equation} \label{DrampDa}
\frac{\partial{}}{\partial{}a_k}L(y_i,F(\mathbf{x}_i))=-2(y_i-F(\mathbf{x}_i))\,x_{ik}\,I(|F(\mathbf{x}_i)|<1).
\end{equation}
Using the form of $F(\mathbf{x}_i)$ in the partial derivative (\ref{DrampDa}) and then substituting into (\ref{gradient}), we get the gradient for the risk using the squared error ramp loss function
\begin{equation} \label{negativegradient}
\begin{array}{r l}
g_k & = \displaystyle \frac{2}{N}\sum_{i=1}^N\,\left(y_i - a_0 - \sum_{j=1}^K\,a_j\,x_{ij}\right)x_{ik}\,I(|F(\mathbf{x}_i)|<1).\\ \\
\end{array} 
\end{equation}
Rearranging, switching the order of summation, and evaluating at the $\ell$-th step $\mathbf{a}^\ell$ in the approximation of $\mathbf{a}$ we can write the gradient at the $\ell$-th step as
\begin{equation} \label{lthnegativegradient}
\begin{array}{r l}
g_k^\ell = & \displaystyle \frac{2}{N} \sum_{i=1}^N\,y_i\,x_{ik}\,I(|F^\ell(\mathbf{x}_i)|<1)\\ \\
 & \displaystyle - a_0\frac{2}{N}\sum_{i=1}^N\,x_{ik}\,I(|F^\ell(\mathbf{x}_i)|<1)\\ \\
 & \displaystyle - \sum_{j=1}^K\,a_j^\ell\left[\frac{2}{N}\sum_{i=1}^N\,x_{ij}\,x_{ik}\,I(|F^\ell(\mathbf{x}_i)|<1)\right].
\end{array} 
\end{equation}

\subsection{Negative gradient with auxiliary functions $v$, $u$}

We need to keep track of the dependencies and update properly at each iteration. The goal of the method is to update the coefficients $\mathbf{a}$. We take a step with respect to $\mathbf{a}$ and then update everything, so let $\mathbf{a}$ act as the independent variable. Recall that $i$ is the index over the observations so $x_i$ is the attribute values for the $i$-th observation and $F_i$ is the predicted value for that observation. This leaves us with 
\begin{equation*} \begin{array}{l}
F^\ell(\mathbf{x}_i) =a_0 + \displaystyle \sum_{k = 1} ^{K} a^\ell_k\, x_{ik} = F_i(a^\ell).
\end{array} \end{equation*}

Defining the indicators $v_i^\ell$
\begin{equation*} \begin{array}{l}
v^\ell_i = v_i(a^\ell) = I( |F_i(a^\ell)|<1),\\ \\
\end{array} \end{equation*}

we can define a new function by
\begin{equation}
\label{u_ramp}
u^\ell(p,q) = u(v^\ell; p,q) = \frac{2}{N}\sum_{i=1}^N\,p_i\,q_i\,v_i^\ell
\end{equation}
where $p$ and $q$ are scalars and $v_i^\ell=I(|F^\ell(\mathbf{x}_i)|<1)$. Using the two functions $v, u$ the negative gradient at the $\ell$-th step (\ref{lthnegativegradient}) can be written in a simpler form
\begin{equation} \label{friedmangrad}
g_k^\ell = u^\ell(v^\ell; y,x_k)- a_0\, u^\ell(v^\ell; 1,x_k) - \sum_{j=1}^K\,a_j^\ell\,u^\ell(v^\ell; x_j,x_k).
\end{equation}

\section{Fast Algorithm}
To ``step" we move proportional to the largest component of the negative gradient (\ref{gradient}). Let $g^\ell_{j^*}$ be the largest absolute component of the gradient 
$$ j^*=\arg\max_{1\leq j\leq K}|g_j^\ell|$$
at the $\ell$-th step. Then call the length of the next step $\delta_* = \Delta_\nu\,g_{j^*}^\ell$ and update the coefficients with $a_{j^*}^{\ell+1}=a_{j^*}^\ell+ \delta_*$. The coefficients at the $( \ell + 1)$-th step are 
\begin{equation}\label{aupdate}
\mathbf{a}^{\ell+1}= \begin{cases} 
a_{j}^\ell, & \mathrm{if}\, j = 1..K, j \neq j^* \\
a_{j^*}^\ell+ \delta_* & \mathrm{if}\, j = j^*.\\
\end{cases}
\end{equation}
After a step the gradient must be recomputed before another step can be taken. Rather than fully recomputing an update can be applied only to the components of the gradient that are affected by the step. There are two cases of how the update to the gradient can be made. One update occurs when the step in the coefficients has caused indicator functions to change; this update requires more work and is expensive. The other update is cheap and is given as follows.

\subsection{Case when indicators do not change}
The step size $\Delta_\nu$ should be small; in practice it is taken to be 0.01. The idea is that with a small stepsize $|F(\mathbf{x}_i)|$ will not exceed 1 ``often." On the steps where this is true the indicators do not change so $v^\ell, u(v^\ell; y,x_k), u(v^\ell; x_j,x_k)$ do not change and the negative gradient at the $( \ell + 1)$-th step is found by substituting (\ref{aupdate}) into (\ref{friedmangrad})

\begin{equation}\label{gradupdate}
\begin{array}{r c l}
g_k^{\ell+1} = & \displaystyle u(v^\ell; y,x_k) - a_0\, u^\ell(v^\ell; 1,x_k) - \sum_{j=1}^K\,a_j^\ell\,u(v^\ell; x_j,x_k) & - \delta_*\,u(v^\ell; x_{j^*},x_k) \\

		= & \displaystyle g_k^\ell & - \delta_*\,u(v^\ell; x_{j^*},x_k).
\end{array}
\end{equation}

\subsection{Case when indicators change - adjustments}
If the assumption fails and the indicators change on a step, then $v^\ell \neq v^ {\ell+1}$ $u^\ell(v^\ell) \neq u^{\ell+1}(v^\ell)$ and (\ref{gradupdate}) does not hold. To find $g^{\ell+1}$, consider the cases of how $v$ can change and and define the variable
\begin{equation}
\label{z}
z_i^\ell=
\begin{cases}
-1, & \mathrm{if}\;v_i^{\ell}=1\; \text{and} \;v_i^{\ell + 1}=0\\
\;\;\,0, & \mathrm{if}\;v_i^\ell=v_i^{\ell + 1}\\
+1, & \mathrm{if}\;v_i^\ell=0\; \text{and} \;v_i^{\ell + 1}=1.
\end{cases}
\end{equation}
$z_n$ can be thought of adding in observations where the indicators have turned on and subtracting observations where indicators have turned off. Using $z_n$, $u$ can be adjusted 
\begin{equation}\label{uupdate}
\begin{array}{r l l}
u(v^{\ell + 1}; y,x_k) = & u(v^\ell; y, x_k) & + \displaystyle \frac{2}{N} \sum_{z_n \neq 0}z_n\,y_n\,x_{nk}\\
u(v^{\ell + 1} ;x_j,x_k) = & u(v^\ell; x_j, x_k) & + \displaystyle \frac{2}{N} \sum_{z_n \neq 0}z_n\,x_{nj}\,x_{nk}\\
u(v^{\ell + 1} ;1,x_k) = & u^\ell(v^\ell; 1,x_k) & + \displaystyle \frac{2}{N} \sum_{z_n \neq 0}z_n\,x_{nk}\\
\end{array}
\end{equation}

and used with (\ref{aupdate}) and (\ref{negativegradient}) to find the $\ell$-th update of the negative gradient 
\begin{equation}\label{}
\begin{array}{r c c l}
g_k^{\ell+1} = & \displaystyle u(v^{\ell+1}; y,x_k) &- a_0\,u(v^{\ell+1}; 1, x_k) & - \displaystyle \sum_{j=1}^K\,a_j^{\ell+1}\,u(v^{\ell+1}; x_j,x_k) \\ 
= & \displaystyle u(v^\ell; y, x_k) & - a_0\, u(v^\ell; 1,x_k) & - \displaystyle \sum_{j=1}^K\,a_j^{\ell}\,u(v^{\ell+1}; x_j,x_k)\\
& &+ \displaystyle \frac{2}{N} \sum_{z_n \neq 0}z_ny_nx_{nk} & \displaystyle - a_0\,\frac{2}{N} \sum_{z_n \neq 0}z_nx_{nk} - \delta_*\,u(v^{\ell+1}; x_{j^*},x_k) 
\end{array}
\end{equation}

With a little more rearrangement the update to the gradient as 
\begin{equation}
\begin{array}{r l l}
 g_k^{\ell+1} = 
 & g_k^{\ell} & \text{adjust for obs. with changed $I(|F_i|<1)$}\\ \\
 & + \displaystyle \frac{2}{N} \sum_{z_n \neq 0} z_ny_{n}x_{nk} & \text{update $u^l(y,x_k)$ from indicator change}\\
 & - \displaystyle a_0\,\frac{2}{N} \sum_{z_n \neq 0}z_nx_{nk} & \text{update $u(v^{\ell+1}; 1, x_k)$ from indicator change}\\ 
 & - \displaystyle \delta_*u(v^{\ell}x_{j^*},x_k) & \text{step in $j^*$ direction}\\ \\
 & -\displaystyle \delta_* \frac{2}{N} \sum_{z_n \neq 0} z_nx_{nj^*}x_{nk} & \text{terms not included in update due to old } v^\ell\\
 & - \displaystyle \frac{2}{N} \sum_{j=1}^K\,a_j^{\ell} \sum_{z_n \neq 0} z_n x_{nj^*}x_{nk} & \text{adjust for observations with changed $I(|F_i|<1).$}
 \end{array}
 \end{equation}

\bibliographystyle{plain}
\bibliography{RuleEnsembleBibliography}{}
\end{document}